\documentclass[manuscript,screen]{acmart}

\AtBeginDocument{%
  }

\definecolor{mygray}{gray}{.85}
\definecolor{mygray1}{gray}{.7}
\definecolor{mygray2}{gray}{.93}
\definecolor{myblue}{HTML}{7F7FFF} 
\definecolor{myred}{HTML}{E95332}
\definecolor{mygreen}{HTML}{199524}
\definecolor{mysec1}{HTML}{000000}
\definecolor{mysec2}{HTML}{219EBC}
\definecolor{mysec3}{HTML}{54426B}
\definecolor{mysec4}{HTML}{215968}
\definecolor{mysec5}{HTML}{936639}
\definecolor{mysec6}{HTML}{6A994E}
\definecolor{mysec7}{HTML}{EE9B00}
\definecolor{mysec8}{HTML}{000000}

\usepackage{graphicx}        
\usepackage{array, booktabs} 
\usepackage{multirow}        
\usepackage{xcolor}   
\usepackage{colortbl}        
\usepackage{arydshln}        
\usepackage{bm}
\usepackage{xspace}
\usepackage{algorithm}
\usepackage{listings}
\usepackage{etoolbox}
\makeatletter
\AfterEndEnvironment{algorithm}{\let\@algcomment\relax}
\AtEndEnvironment{algorithm}{\kern2pt\hrule\relax\vskip3pt\@algcomment}
\let\@algcomment\relax
\newcommand\algcomment[1]{\def\@algcomment{\footnotesize#1}}
\renewcommand\fs@ruled{\def\@fs@cfont{\bfseries}\let\@fs@capt\floatc@ruled
  \def\@fs@pre{\hrule height.8pt depth0pt \kern2pt}
  \def\@fs@post{}
  \def\@fs@mid{\kern2pt\hrule\kern2pt}
  \let\@fs@iftopcapt\iftrue}
\makeatother

\usepackage{tabulary}
\newcolumntype{I}{!{\vrule width 1pt}}
\newcolumntype{x}[1]{>{\centering\arraybackslash}p{#1pt}}
\newcolumntype{y}[1]{>{\raggedright\arraybackslash}p{#1pt}}
\newcolumntype{z}[1]{>{\raggedleft\arraybackslash}p{#1pt}}

\definecolor{codegreen}{RGB}{79,126,127}
\definecolor{codedefine}{RGB}{153,54,159}
\definecolor{codefunc}{RGB}{73,122,234}
\definecolor{codecall}{RGB}{73,122,234}
\definecolor{codepro}{RGB}{212,96,80}
\definecolor{codedim}{RGB}{89,152,195}
\definecolor{3dgc1}{RGB}{177, 83, 74}
\definecolor{3dgc2}{RGB}{93, 107, 72}

\makeatletter
\newcommand{\thickhline}{
    \noalign {\ifnum 0=`}\fi \hrule height 1pt
    \futurelet \reserved@a \@xhline
}
\makeatother

\makeatletter
\DeclareRobustCommand\onedot{\futurelet\@let@token\@onedot}
\def\@onedot{\ifx\@let@token.\else.\null\fi\xspace}

\def\eg{\emph{e.g}\onedot} 
\def\ie{\emph{i.e}\onedot}

\makeatother

\begin{document}

\title[WMAD]{A Survey of World Models for Autonomous Driving}

\author{Tuo Feng}
\email{fengtuo2015@outlook.com}
\orcid{0000-0001-5882-3315}
\author{Wenguan Wang}
\authornote{Corresponding author}
\email{wenguanwang.ai@gmail.com}
\orcid{0000-0002-0802-9567}
\author{Yi Yang}
\email{yangyics@zju.edu.cn}
\orcid{0000-0002-0512-880X}
\affiliation{%
  \institution{Collaborative Innovation Center of Artificial Intelligence (CCAI), Zhejiang University}
  \city{Hangzhou}
  \state{Zhejiang}
  \country{China}
}

\renewcommand{\shortauthors}{T. Feng et al.}

\begin{abstract}
Recent breakthroughs in autonomous driving have been propelled by advances in robust world modeling, fundamentally transforming how vehicles interpret dynamic scenes and execute safe decision-making. World models have emerged as a linchpin technology, offering high-fidelity representations of the driving environment that integrate multi-sensor data, semantic cues, and temporal dynamics. This paper systematically reviews recent advances in world models for autonomous driving, proposing a three-tiered taxonomy: (i) Generation of Future Physical World, covering Image-, BEV-, OG-, and PC-based generation methods that enhance scene evolution modeling through diffusion models and 4D occupancy forecasting; (ii) Behavior Planning for Intelligent Agents, combining rule-driven and learning-based paradigms with cost map optimization and reinforcement learning for trajectory generation in complex traffic conditions; (ii) Interaction between Prediction and Planning, achieving multi-agent collaborative decision-making through latent space diffusion and memory-augmented architectures. The study further analyzes training paradigms, including self-supervised learning, multimodal pretraining, and generative data augmentation, while evaluating world models' performance in scene understanding and motion prediction tasks. Future research must address key challenges in self-supervised representation learning, multimodal fusion, and advanced simulation to advance the practical deployment of world models in complex urban environments. Overall, the comprehensive analysis provides a technical roadmap for harnessing the transformative potential of world models in advancing safe and reliable autonomous driving solutions.
\end{abstract}



\received{7 September 2025}

\maketitle

\section{Introduction}\label{sec:intro}

The quest for autonomous driving has become a focal point in both scientific research and industry endeavors. At its core lies the ambition to reduce traffic accidents, alleviate congestion, and enhance mobility for societal groups~\cite{fagnant2015preparing}. Current statistics underscore that human error remains the principal cause of accidents~\cite{world2019global}, indicating that reducing reliance on direct human control could lower the incidence of traffic-related fatalities and injuries. Beyond safety, economic factors also propel the development of autonomous driving technologies~\cite{Liu2020Computing}.

Despite these incentives, achieving high-level autonomy faces major hurdles. Foremost among these is perceiving and understanding dynamic traffic scenarios, which requires fusing heterogeneous sensor streams into an environmental representation~\cite{Grigorescu2019A, Furda2011Enabling}. Autonomous vehicles must rapidly assimilate multi-modal data, detect salient objects, and anticipate their motion under adversarial conditions -- such as sensor degradation in heavy rain, unpaved roads with poor markings, or aggressive traffic maneuvers~\cite{schwarting2018planning, manivasagam2020lidarsim}. Moreover, real-time decision-making introduces computational constraints, imposing millisecond-level responsiveness to address unexpected obstacles or anomalous behaviors~\cite{Li2023Research}. Equally pivotal is the system's resilience in extreme or long-tail scenarios (\eg, severe weather, construction zones, or erratic driving behaviors), where performance shortfalls can compromise overall safety~\cite{bogdoll2024umad, wang2024drivedreamer}.

Within this context, constructing robust \textit{world models} has emerged as a cornerstone. In essence, a world model is a generative spatio-temporal neural system that compresses multi-sensor physical observations into a compact latent state and rolls it forward under hypothetical actions, letting the vehicle rehearse futures before they occur~\cite{ding2024understanding,ha2018world}. This high-fidelity internal map~\cite{Liu2020Computing} updates online, and feeds downstream tasks such as physical-world prediction~\cite{Qian2019Deep}. Recent work refines such models with generative sensor fusion~\cite{li2024drivingdiffusion, hu2023gaia, hassan2024gem, bian2025dynamiccity, wang2024occsora}, unifying heterogeneous inputs into consistent top-down views~\cite{Wang2020Research, yang2024visual}.

These robust world models leverage environmental representations to optimize behavior planning, serving as a keystone for safer, efficient autonomous driving. By enabling trajectory optimization, real-time hazard detection, and adaptive route planning, they reduce predictable risks~\cite{Furda2011Enabling} and are compatible with vehicle-to-everything systems~\cite{Li2023Research}. World models facilitate cohesive integration between perception and control subsystems, streamlining the closed-loop autonomy pipeline~\cite{min2024driveworld, zheng2025genad}.

\begin{figure*}[!t]
\centering
\includegraphics[width=0.99\textwidth]{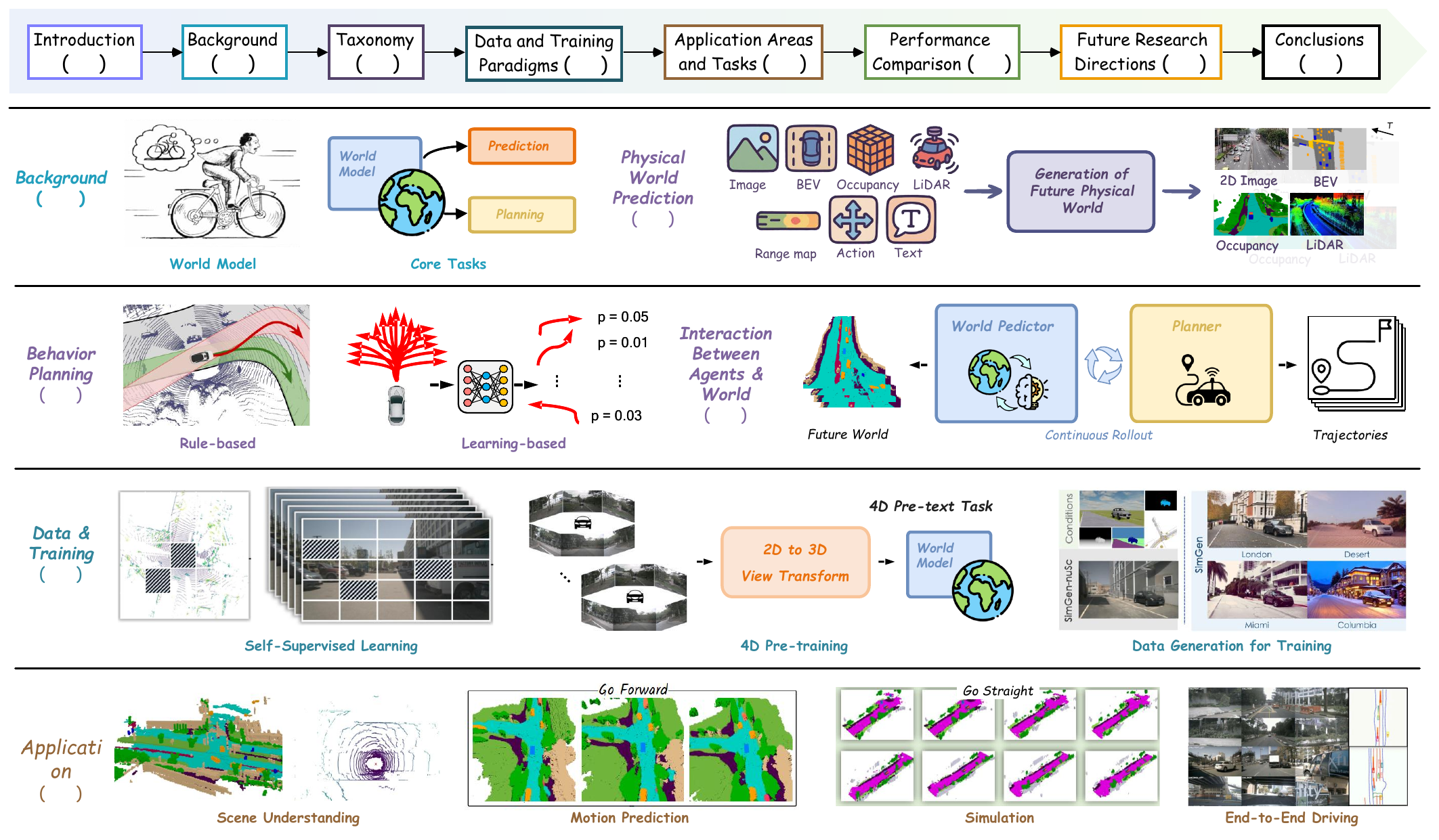}
\put(-404, 228){\scriptsize{\S\ref{sec:intro}}}
\put(-359, 228){\scriptsize{\S\ref{sec:background}}}
\put(-316, 228){\scriptsize{\S\ref{sec:Taxonomy}}}
\put(-256, 228){\scriptsize{\S\ref{sec:Data_and_Training_Paradigms}}}
\put(-196, 228){\scriptsize{\S\ref{sec:application}}}
\put(-135, 228){\scriptsize{\S\ref{sec:performance}}}
\put(-78, 228){\scriptsize{\S\ref{sec:future_directions}}}
\put(-38, 228){\scriptsize{\S\ref{sec:conclusion}}}
\put(-410, 188){\scriptsize{\S\ref{sec:background}}}
\put(-237, 182){\scriptsize{\S\ref{sec:methodologies_future_prediction}}}
\put(-412.5, 129.5){\scriptsize{\S\ref{sec:methodologies_behavior_planning}}}
\put(-216.3, 123.9){\scriptsize{\S\ref{sec:methodologies_interaction}}}
\put(-411, 77){\scriptsize{\S\ref{sec:Data_and_Training_Paradigms}}}
\put(-411, 12.5){\scriptsize{\S\ref{sec:application}}}
\caption{\small \textbf{Structure of the overall review} (\S\ref{sec:intro}). The top row outlines the organization. The second and third rows illustrate the background and key components. The fourth row highlights various methodologies for training models in autonomous driving. The bottom row showcases four application areas for world models in autonomous driving.}
\label{fig_structure}
\vspace{-12pt} 
\end{figure*}

\noindent\textbf{Differences between Our Survey and Others.} Existing surveys on world models for autonomous driving can be classified into two categories. The mainstream category covers general world models used across many fields~\cite{zhu2024sora,ding2024understanding}, with autonomous driving only one area. The second category~\cite{guan2024world, fu2024exploring}, though relatively scarce, focuses on the application of world models within autonomous driving. They categorize studies coarsely and often focus solely on world simulation or lack discussions on the interaction between planning and prediction, resulting in a lack of a clear taxonomy. In this paper, a comprehensive review of world models for autonomous driving is presented, and their applications are explored, thereby highlighting how these models are shaped by and adapt to the automotive sector.

\noindent\textbf{Goals of Our Survey.} Guided by the principle that world models are central to the understanding of dynamic scenes, this survey aims to (i) develop a \textit{comprehensive, structured taxonomy} of world models for autonomous driving highlighting the core architectures, input/output modalities, and conditioning strategies; (ii) classify the most commonly used datasets and evaluation metrics employed in world‐model research for autonomous driving; and (iii) identify future research directions to advance the reliability and safety of world models in complex driving environments.

Current research falls into three key areas: (i) Generation of Future Physical World: Focusing on the physical world evolution of both dynamic objects and static entities~\cite{wang2024drivedreamer, zhao2024drivedreamer4d, wang2024occsora}; (ii) Behavior Planning for Intelligent Agents: Examining generative and rule-based planning methods that produce safe, efficient paths under uncertain driving conditions~\cite{Qian2019Deep,hu2022st,casas2021mp3}; (iii) Interaction between Behavior Planning and Future Prediction: Highlighting how unified frameworks can capture agent interactions and leverage predictive insights for collaborative optimization~\cite{yang2024drivearena,yang2024drivearena}. Specifically, the following contributions are presented:
\vspace{-6pt} 
\begin{itemize}
    \item Analysis of Future Prediction Models: Image-/BEV-/OG-/PC-based generation methods are examined to show how geometric and semantic fidelity is achieved, including 4D occupancy forecasting and diffusion-based generation.
    \item Investigation of Behavior Planning: Behavior planning is explored through both rule-based and learning-based approaches, and notable improvements in robustness and collision avoidance are demonstrated.
    \item Proposition of Interactive Model Research: Interactive models that jointly address future prediction and agent behavior are systematically reviewed, indicating how this synergy can vastly enhance real-world adaptability and operational safety.
\vspace{-6pt} 
\end{itemize}

Four frontier challenges are highlighted -- developing \textit{self-supervised} world models, constructing \textit{unified multi-modal} embeddings, building \textit{advanced, physics-aware} simulators, and designing \textit{lean, latency-aware} architectures -- to chart a clear roadmap for next-generation autonomous driving research. As the research landscape expands and real-world adoption grows more urgent, this survey is intended to serve as a valuable reference for researchers and practitioners, laying the groundwork for safer, more robust autonomous-driving solutions.

A summary of this paper's structure is given in Fig.~\ref{fig_structure}: \S\ref{sec:intro} motivates world models for autonomous driving; \S\ref{sec:background} formalises the core tasks of world models; \S\ref{sec:Taxonomy} introduces a taxonomy of existing methods; \S\ref{sec:Data_and_Training_Paradigms} contrasts prevailing data and training paradigms; \S\ref{sec:application} outlines key application areas; \S\ref{sec:performance} benchmarks state-of-the-art models; \S\ref{sec:future_directions} identifies open challenges and research opportunities; and \S\ref{sec:conclusion} concludes the survey and summarizes key findings.

\section{Background}
\label{sec:background}

\S\ref{sec:background_context} first defines world models and their autonomous driving variant that fuses multi-modal inputs for long-horizon planning tasks; next, \S\ref{sec:background_formulation} states the formal problem, describing future-world generation and agent planning.

\subsection{Definition of World Models}
\label{sec:background_context}

A world model is a generative spatio-temporal neural system that encodes the external physical environment into a compact latent state, jointly capturing geometry, semantics and causal context~\cite{ding2024understanding}. This internal state is learned without labels: the system first employs a self-learning compressor to squeeze raw sensor frames into a handful of key numbers; next, a time-aware prediction module uses the hidden state and the agent's action to infer what the next hidden state will be, allowing the agent to rehearse an entire trajectory in its ``mind'' before acting in the real world~\cite{ha2018recurrent,ha2018world}. Because every part of the pipeline is differentiable~\cite{ha2018world}, the whole model functions as an interpretable virtual sandbox that supplies gradients, data samples, and ``what-if'' roll-outs, enabling the agent to make decisions more efficiently~\cite{ding2024understanding}.

World Models for Autonomous Driving apply the same idea to road domains by mapping synchronized camera images, LiDAR sweeps, radar echoes and HD-maps into a single latent scene graph, thereby unifying perception and prediction within one representation~\cite{guan2024world}. Inside this latent space the model rolls forward the joint dynamics of surrounding traffic and the ego vehicle over long horizons~\cite{ding2024understanding}, enabling realistic scene generation, multi-agent behaviour forecasting, trajectory evaluation and low-latency control -- all while respecting the physical constraints and multimodal uncertainties endemic to real-world driving.

\subsection{Core Tasks in World Models for Autonomous Driving\!\!\!}
\label{sec:background_formulation}

In autonomous driving, a critical aspect is accurately predicting the future states of both the ego vehicle and its surrounding environment. To address the core tasks, world models $\bm{w}$ in autonomous driving takes sensor inputs (including a set of multi-view images $\bm{I}$ and a set of LiDAR points $\bm{P}$) collected from previous frames and infers the scene and trajectory for the next frames. Specifically, the ego trajectory at time $T+1$, denoted as $\tau^{T+1}$, is predicted alongside the surrounding scene $\bm{z}^{T+1}$. $\bm{w}$ models the coupled dynamics of the ego vehicle's motion and the environment's evolution. Formally, the function $\bm{w}$ is given by:
\begin{equation}\label{eqn:model}
\begin{aligned}
\!\!\!\!\bm{z}^{T+1},\! \tau^{T+1}\! =\! \bm{w}((\bm{I}^T,\! \cdots,\! \bm{I}^{T-t}),\! (\bm{P}^T,\! \cdots,\! \bm{P}^{T-t})).
\end{aligned}
\end{equation}

The first core task is generation of future physical world~\cite{agro2024uno,zhang2024bevworld,yang2024visual,hu2021fiery}, which involves forecasting the future states of dynamic entities. This task emphasizes capturing potential interactions, stochastic behaviors, and uncertainties within rapidly changing and complex scenes. The second core task is behavior planning for intelligent agents~\cite{hu2022st,hu2022model,zheng2025genad,popov2024mitigating}, focusing on generating optimal and feasible trajectories for the ego vehicle. It requires accounting for safety constraints, dynamic obstacles, traffic regulations, and real-time adaptability.

\section{Taxonomy}
\label{sec:Taxonomy}

\subsection{Generation of Future Physical World}
\label{sec:methodologies_future_prediction}

\begin{figure}[t]
\centering
\includegraphics[width=0.89\linewidth]{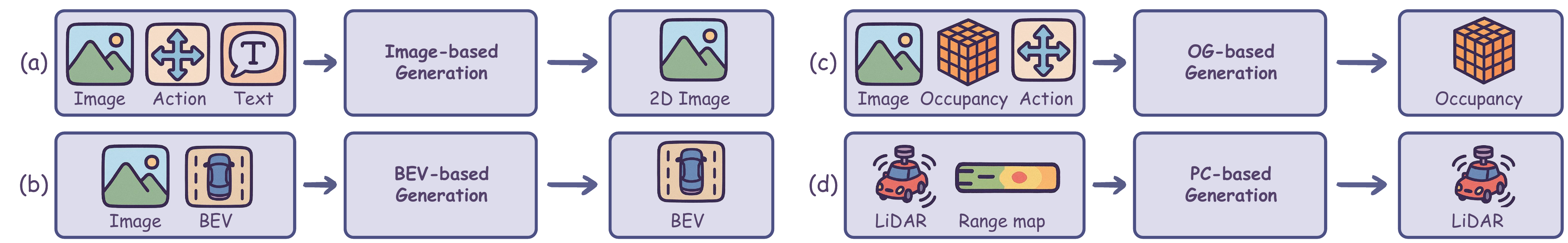}
\caption{\small \textbf{The paradigms of generation methods for future physical world} (\S\ref{sec:methodologies_future_prediction}). (a) Image-based generation synthesizes high-fidelity 2D images. (b) BEV-based generation forecasts BEV maps using paired image and BEV cues. (c) OG-based generation predicts 4D occupancy grids. (d) PC-based generation outputs future LiDAR sweeps. Boxes indicate processing modules; solid arrows denote data flow.}
\label{fig2}
\vspace{-6pt} 
\end{figure}

Fig.~\ref{fig2} arranges future-scene generators into four tracks -- image, Bird's-Eye View (BEV), occupancy grid (OG) and point cloud (PC) -- that jointly raise realism, from photoreal frames through map-level layouts to 4D voxels and LiDAR sweeps, all enabled by controllable diffusion and long-horizon forecasting.

\subsubsection{Image-based Generation}
\label{sec:Image-based}

Image-based future-prediction methods for autonomous driving use generative models to tackle limited data and dynamic environments. Synthesizing realistic images enlarges training data and strengthens perception and planning.

\noindent\textbf{Dreamer Series.} \textit{Dreamer series} was first proposed for learning world models to address dynamic scene understanding in gaming and robotics~\cite{hafnerdream,hafnermastering,hafner2023mastering}, later expanded to autonomous driving with DriveDreamer~\cite{wang2024drivedreamer}. It enables controllable video generation in high-density urban traffic using real-world driving data, though they remain limited to 2D outputs lacking spatial-temporal coherence. Subsequent research introduces DriveDreamer-2~\cite{zhao2025drivedreamer}, incorporating LLM-based prompting mechanisms to enhance interactivity and diversity. With further advancements, DriveDreamer4D~\cite{zhao2024drivedreamer4d} significantly extends model capabilities by utilizing world model priors to generate spatial-temporally coherent 4D driving videos, marking a transition from traditional 2D to advanced 4D video generation. ReconDreamer~\cite{ni2024recondreamer} explores online restoration techniques to achieve precise video reconstruction of dynamic scenes. WorldDreamer~\cite{wang2024worlddreamer} expands the applicability of world models by pioneering masked token prediction for multi-modal generation, bridging text-to-video and action-to-video tasks, and moving towards general-purpose world model construction. The development of the \textit{Dreamer series} demonstrates a clear trend: progressing from 2D to multi-dimensional video generation, transitioning from task-specific to general-purpose models, and shifting from closed, structured conditions toward more open and general natural-language-based conditions.

\noindent\textbf{Diffusion-based Image Generation.} The evolution of this type in autonomous driving are coalescing around two complementary thrusts: \textit{controllable generation} (\eg, BEVControl~\cite{yang2023bevcontrol}, DrivingDiffusion~\cite{li2024drivingdiffusion}, and GeoDrive~\cite{chen2025geodrive}), which use latent diffusion to render geometry-consistent videos from rich cues such as BEV layouts, text prompts and optical flow, and \textit{high-fidelity spatio-temporal modelling}, where Drive-WM~\cite{wang2024driving}, Vista~\cite{gao2024vista}, and LongDWM~\cite{wang2025longdwm} push diffusion roll-outs to longer horizons and higher spatial resolutions, markedly boosting temporal coherence and per-frame detail. The field progresses through breakthroughs in \textit{multi-modal conditions}: integrating ego trajectory, human-pose, and DINO appearance tokens for behaviour-aware video synthesis~\cite{hassan2024gem}; translating simulator layouts into photorealistic scenes through a cascade diffusion pipeline~\cite{zhou2024simgen} or dual-scale controlNet-stable diffusion~\cite{li2025simworld}; uniting LiDAR and camera streams in a BEV latent grid for geometry-consistent cross-sensor forecasting~\cite{zhang2024bevworld}; aligning ego–other vehicle trajectories in latent space to enable fully controllable, multi-agent video generation~\cite{zhu2025other}; and grounding multi-camera driving video generation in physics by aligning ego-world coordinates, injecting 3D flows, and box-guided occlusion reasoning~\cite{yang2024pysical}. Recent methods~\cite{liang2025unifuture, li2025driverse, jiang2025dive, ji2025cogen, russell2025gaia, wang2025mila, ma2024unleashing,wang2025stage} focus on expanding from simple RGB image generation to more \textit{diverse and enriched paradigms}, including extending conventional RGB prediction to jointly generate both images and depth~\cite{liang2025unifuture}; shifting from layout-conditioned synthesis to fine-grained trajectory-conditioned control~\cite{li2025driverse}; moving beyond 2D-only conditioning toward multi-view consistent generation~\cite{jiang2025dive}; transitioning from 2D layout cues to full 3D scene conditioned synthesis~\cite{ji2025cogen}; broadening single-view synthesis into multi-view, multi-agent interactive scenarios~\cite{russell2025gaia}; progressing from short-term clips to long-horizon video generation~\cite{wang2025mila}; and further leveraging failure-case augmentation for robustness~\cite{ma2024unleashing}; alongside the release of accident-anticipation benchmarks for risk-aware evaluation~\cite{guan2025world}. Moreover, InfiniCube~\cite{lu2024infinicube} employs world-guided video diffusion grounded on HD maps, 3D bounding boxes, and text, then lifts the generated long-horizon videos into dynamic 3D Gaussian scenes via a fast feed-forward reconstruction pipeline. This technical evolution reveals a clear trend from isolated scene generation to integrated closed-loop simulation systems, providing high-fidelity, scalable synthetic environments for autonomous driving.

\noindent\textbf{Transformer-based Image Generation.} Transformer-based driving imagers are evolving from \textit{narrow, single-view renderers} into \textit{holistic, action-aware world models}. \textit{Cross-view} architectures such as HoloDrive~\cite{wu2024holodrive} and BEVGen~\cite{swerdlow2024street} fuse camera and LiDAR or lift BEV layouts to street-level frames, unifying 2D appearance with 3D geometry. A \textit{parallel line} scales sequence modeling: DrivingWorld~\cite{hu2024drivingworld} and GAIA-1~\cite{hu2023gaia} treat video, text and control as one token stream, enabling minute-long, controllable clips with coherent semantics. These token factories prove that unsupervised prediction alone can encode road rules and actor intent. \textit{Complementary work} probes limits and remedies: DriveSim~\cite{sreeram2024probing} exposes causal hallucinations in multimodal LLMs, while a DINOv2-based multi-view encoder~\cite{popov2024mitigating} compresses synchronized front views into a latent scene grid that decodes to both RGB and BEV semantics, reducing covariate shift.

\begin{table*}[t]
\centering
\caption{\small \textbf{Overview} of summary, methods and core architectures for Image-/BEV-based generation (\S\ref{sec:Image-based} and \S\ref{sec:bev-based}).}
    \small
    \resizebox{0.99\textwidth}{!}{%
      \setlength\tabcolsep{0.8pt}
      \renewcommand\arraystretch{1.05}
\begin{tabular}{r||ccccccc}
\hline\thickhline\rowcolor{mygray}
\textbf{Summary} & \textbf{Method}& \textbf{Pub.} & \textbf{Core Architecture} & \textbf{Input Modality} \& \textbf{Control Condition} & \textbf{Output Modality} & \textbf{Training Dataset}\\
\hline\hline
\multicolumn{7}{c}{\textit{Image-based Generation}}\\
\hline
\multirow{6}{*}{\shortstack{Image-based Generation\\Dreamer Series}}
& DriveDreamer\cite{wang2024drivedreamer} & ECCV'24 & Diffusion Model & Text\,{+}\,Image\,{+}\,HDMap\,{+}\,Box\,{+}\,Actions & 2D Image\,{+}\,Actions & nuScenes\cite{caesar2020nuscenes} \\
& DriveDreamer-2\cite{zhao2025drivedreamer} &AAAI'25 & LLM\,{+}\,Diffusion Model & Text\,{+}\,HDMap\,{+}\,Box & 2D Image & nuScenes\cite{caesar2020nuscenes}\\
& DriveDreamer4D\cite{zhao2024drivedreamer4d} & CVPR'24 & Diffusion Model\,{+}\,4DGS &Text\,{+}\,Image\,{+}\,Trajectory& 2D Image& Waymo\cite{sun2020scalability} \\
& ReconDreamer\cite{ni2024recondreamer} & CVPR'24  &Diffusion Model & Image\,{+}\,HDMap\,{+}\,Box &2D Image & Waymo\cite{sun2020scalability}\\
& WorldDreamer\cite{wang2024worlddreamer} & arXiv'24 & LLM & Image\,{+}\,Video\,{+}\,Text\,{+}\,Action & 2D Image&nuScenes\cite{caesar2020nuscenes}\\
& CarDreamer\cite{gao2024cardreamer} &IOTJ'24 &World Model Backbone & Image\,{+}\,BEV\,{+}\,LiDAR\,{+}\,Radar&  2D Image\,{+}\,Actions& CARLA\cite{dosovitskiy2017carla}\\
 \cdashline{1-7}[1pt/1pt]
\multirow{24}{*}{\shortstack{Image-based Generation\\Diffusion Models}} & BEVControl~\cite{yang2023bevcontrol} & arXiv'23 & Diffusion Model & BEV\,{+}\,Image\,{+}\,Text & 2D Image & nuScenes\cite{caesar2020nuscenes}\\
& DrivingDiffusion\cite{li2024drivingdiffusion} & ECCV'24 & Diffusion Model & Image\,{+}\,Layouts\,{+}\,Text\,{+}\,Optical Flow &2D Image & CARLA\cite{dosovitskiy2017carla}\\
& Drive-WM\cite{wang2024driving} & CVPR'24 & Diffusion Model & Image\,{+}\,Action\,{+}\,Box\,{+}\,Map\,{+}\,Text & 2D Image & nuScenes\cite{caesar2020nuscenes}\,{+}\,Waymo\cite{sun2020scalability}\\
& DriVerse~\cite{li2025driverse} & ACMMM'25 & Diffusion Model & Image\,{+}\,Trajectory & 2D Image & nuScenes\cite{caesar2020nuscenes}\,{+}\,Waymo\cite{sun2020scalability}\\
& GEM\cite{hassan2024gem} & CVPR'24 & Diffusion Model & Image\,{+}\,Trajectory\,{+}\,Human pose\,{+}\,DINO features & 2D Image\,{+}\,Depth & Self-created dataset\cite{hassan2024gem} \\
& Vista\cite{gao2024vista}& NeurIPS'24 & Diffusion Model & Image\,{+}\,Action & 2D Image & Self-created dataset\cite{gao2024vista} \\
& SimGen~\cite{zhou2024simgen} & NeurIPS'24 & Diffusion Model & Image\,{+}\,Depth\,{+}\,Segmentation\,{+}\,Text & 2D Image& DIVA~\cite{zhou2024simgen}\,{+}\,nuScenes\cite{caesar2020nuscenes} \\
& Delphi\cite{ma2024unleashing} & arXiv'24 & Diffusion Model  & Image\,{+}\,BEV layout\,{+}\,Failure Case &2D Image&nuScenes\cite{caesar2020nuscenes}\\
& BevWorld\cite{zhang2024bevworld}& arXiv'24 & Diffusion Model & Image\,{+}\,Point cloud\,{+}\,Action & 2D Image\,{+}\,3D LiDAR & nuScenes\cite{caesar2020nuscenes}\,{+}\,CARLA\cite{Dosovitskiy17} \\
& DrivePhysica~\cite{yang2024pysical} & arXiv'24 & Diffusion Model & Image\,{+}\,Poses\,{+}\,Box\,{+}\,Lane\,{+}\,Text & 2D Image & nuScenes\cite{caesar2020nuscenes}\\
& Imagine-2-Drive~\cite{garg2024imagine} & arXiv'24 & Diffusion Model & Image & 2D Image& CARLA\cite{Dosovitskiy17} \\
& MaskGWM\cite{ni2025maskgwm} & arXiv'25 & Diffusion Model & Image\,{+}\,Text\,{+}\,Action  & 2D Image & OpenDV-2K\cite{yang2024generalized}\,{+}\,nuScenes\cite{caesar2020nuscenes}\\
& EOTWM~\cite{zhu2025other} & arXiv'25 & Diffusion Model & Image\,{+}\,Text\,{+}\,Trajectory  & 2D Image & nuScenes\cite{caesar2020nuscenes}\\
& UniFuture~\cite{liang2025unifuture} & arXiv'25 & Diffusion Model & Image\,{+}\,Depth  & 2D Image\,{+}\,Depth & nuScenes\cite{caesar2020nuscenes}\\
& DiVE~\cite{jiang2025dive} & arXiv'25 & Diffusion Model & Image\,{+}\,Text\,{+}\,Box\,{+}\,Road condition & 2D Image & nuScenes\cite{caesar2020nuscenes} \\ 
& CoGen~\cite{ji2025cogen} & arXiv'25 & Diffusion Model & Image\,{+}\,Occupancy & 2D Image & nuScenes\cite{caesar2020nuscenes} \\ 
& GAIA-2~\cite{russell2025gaia}  & arXiv'25 & Diffusion Model & Image\,{+}\,Action\,{+}\,Text & 2D Image & Internal dataset\cite{russell2025gaia} \\   
& MiLA~\cite{wang2025mila} & arXiv'25 & Diffusion Model & Image\,{+}\,Waypoints\,{+}\,Text\,{+}\,Camera parameters & 2D Image & nuScenes\cite{caesar2020nuscenes}\\
& DiST-4D~\cite{guo2025dist} &  ICCV'25 & Diffusion Model & Image\,{+}\,Trajectory\,{+}\,BEV\,{+}\,Pose\,{+}\,Box & RGBD image & nuScenes\cite{caesar2020nuscenes}\,{+}\,Waymo\cite{sun2020scalability}\\ 
& SimWorld\cite{li2025simworld} & arXiv'25 & Diffusion Model & Mask\,{+}\,Box\,{+}\,Text & 2D Image & AutoMine\cite{li2022automine}\\
& InfiniCube\cite{lu2024infinicube} & ICCV'25 & Diffusion Model\,{+}\,3DGS & HD Map\,{+}\,Bounding boxes\,{+}\,Text & 2D Image\,{+}\,3DGS & Waymo\cite{sun2020scalability}\\
& STAGE\cite{wang2025stage} & arXiv'25 & Diffusion Model & Image\,{+}\,HD Map\,{+}\,3D Box & 2D Image\,{+}\,Long-horizon video & nuScenes\cite{caesar2020nuscenes} \\
& AoTA\cite{guan2025world} & Comm.\ Eng.'25 & Diffusion Model & Image\,{+}\,HD Map\,{+}\,Layouts\,{+}\,Ego pose\,{+}\,Text & 2D Image & nuScenes\cite{caesar2020nuscenes} \\
& GeoDrive~\cite{chen2025geodrive} & arXiv'25 & Diffusion Model & Image\,{+}\,Trajectory & 2D Image & nuScenes\cite{caesar2020nuscenes} \\
& LongDWM~\cite{wang2025longdwm} & arXiv'25 & Diffusion Model & Image\,{+}\,Action & 2D Image\,{+}\,Long-horizon video & nuScenes\cite{caesar2020nuscenes}\\
 \cdashline{1-7}[1pt/1pt]
\multirow{8}{*}{\shortstack{Image-based Generation\\Transformers}}& GAIA-1\cite{hu2023gaia} & arXiv'23 & Transformer & Image\,{+}\,Action\,{+}\,Text & 2D Image & Internal dataset\cite{hu2023gaia}\\   
& HoloDrive\cite{wu2024holodrive} & arXiv'24 & Transformer & Image\,{+}\,LiDAR\,{+}\,Layouts\,{+}\,Text &2D Image\,{+}\,3D LiDAR & nuScenes\cite{caesar2020nuscenes}\\
& BEVGen\cite{swerdlow2024street} & RA-L'24 & Transformer & Image\,{+}\,BEV layout & 2D Image & nuScenes\cite{caesar2020nuscenes}\,{+}\,Argoverse 2\cite{wilson2023argoverse}\\
& DrivingWorld\cite{hu2024drivingworld} & arXiv'24 & Video GPT & Image\,{+}\,Orientation\,{+}\,Location & 2D Image & NuPlan\cite{caesar2021nuplan} Internal data\cite{hu2024drivingworld} \\
& DriveSim~\cite{sreeram2024probing} & arXiv'24 & MLLMs & Image\,{+}\,Text & 2D Image &Internal dataset\cite{sreeram2024probing}\\
& TMPE~\cite{popov2024mitigating} & arXiv'24 & Tansformer &  Image\,{+}\,Path\,{+}\,Speed &  2D Image\,{+}\,BEV & CARLA~\cite{dosovitskiy2017carla}\\
& NeMo\cite{huang2025neural} & ECCV'24 & Transformer & Image & 2D Image\,{+}\,LiDAR\,{+}\,Occupancy & nuScenes\cite{caesar2020nuscenes}\\
& FUTURIS\cite{karypidis2025advancing} & arXiv'25 & Transformer & Segmentation\,{+}\,Depth & Segmentation\,{+}\,Depth & Cityscapes\cite{cordts2016cityscapes} \\ 
\hline\hline
\multicolumn{7}{c}{\textit{BEV-based Generation}}\\
\hline
\multirow{7}{*}{\shortstack{BEV-based Generation}}& FIERY~\cite{hu2021fiery} & ICCV'21 & Transformer  & Image\,{+}\,Ego motion & BEV & nuScenes\cite{caesar2020nuscenes}\\
& StretchBEV\cite{akan2022stretchbev} & ECCV'22 & LSTM & Image & BEV & nuScenes\cite{caesar2020nuscenes}\\
& MILE\cite{hu2022model} &  NeurIPS'22 & RNN  & Image\,{+}\,BEV segmentation\,{+}\,Action & Image\,{+}\,BEV\,{+}\,Action & CARLA\cite{Dosovitskiy17} \\
& UNO\cite{agro2024uno} & CVPR'24 & CNN & LiDAR &  Occupancy\,{+}\,LiDAR\,{+}\,BEV & nuScenes\cite{caesar2020nuscenes}\\
& PowerBEV~\cite{li2023powerbev} & IJCAI'23 & CNN & Image & BEV Instance & nuScenes\cite{caesar2020nuscenes}\\
& GenAD\cite{zheng2025genad} & ECCV'24 & Transformer & Image\,{+}\,BEV & Map\,{+}\,Trajectory & nuScenes\cite{caesar2020nuscenes}\\
&CarFormer\cite{hamdan2025carformer} & ECCV'24 & Transformer & BEV\,{+}\,Trajectory\,{+}\,Destination\,{+}\,Traffic sign & BEV\,{+}\,Action & Longest6\cite{chitta2022transfuser}\\
\hline
\end{tabular}
}
\label{tab:overview1}
\vspace{-12pt} 
\end{table*}

\noindent{\textbf{Summary.} Table~\ref{tab:overview1} shows different modern image-centric generative frameworks, which use diffusion and transformer architectures to produce diverse driving data. Several observations can be drawn:
\begin{itemize}
    \item These methods synthesize photorealistic images to enrich training sets. They also create rare or safety-critical scenarios that are hard to capture in real data.
    \item World models forecast future states conditioned on actions. Fusion of multiple modalities enhances planning and dynamic reasoning.
\vspace{-6pt} 
\end{itemize}

\subsubsection{BEV-based Generation}
\label{sec:bev-based}

\noindent\textbf{BEV Representation.} A BEV representation unifies multi-modal sensor streams into a single top-down map, exposing lane layouts, dynamic actors, and free space in a planner-friendly format~\cite{zhang2024bevworld}. By stripping away perspective distortions, BEV methods provide high-level, spatially coherent context that has become a powerful complement -- or even alternative -- to image-centric pipelines, supporting tasks such as motion prediction and long-horizon trajectory forecasting. This 2D projection, however, gains simplicity at the expense of depth resolution: BEV maps can struggle to retain fine-grained 3D geometry in scenes with complex vertical structure or steep depth gradients~\cite{feng2023clustering}.

Starting with the probabilistic BEV forecaster FIERY~\cite{hu2021fiery}, researchers stretch temporal range with StretchBEV~\cite{akan2022stretchbev}, learn continuous 4D occupancy fields with UnO~\cite{agro2024uno}, and cut computation via the 2DCNN design~\cite{li2023powerbev}. \textit{Object-centric} slots then emerge: CarFormer~\cite{hamdan2025carformer} assigns every actor a latent slot. MILE~\cite{hu2022model} compresses the whole scene into a single compact latent with the driving policy, enabling \textit{policy-aware, multi-modal} prediction. Together, these advances unlock \textit{generative planning}; GenAD~\cite{zheng2025genad} rolls the latent forward to sample ego-conditioned futures, BEVControl~\cite{yang2023bevcontrol} lets users overwrite it with editable sketches for safety auditing, and ViDAR~\cite{yang2024visual}, although operating in point-cloud space, supplies geometry-aware pre-training that lifts BEV performance across the board. Together these works chart a path from early probabilistic forecasts to \textit{efficient, object-aware, user-controllable} world models.

\noindent{\textbf{Summary.} Table~\ref{tab:overview1} summarizes recent BEV‑based generative methods, which predict future BEV representations for autonomous driving tasks. Several observations can be drawn:
\vspace{-4pt} 
\begin{itemize}
\item Early works transform monocular images into probabilistic BEV maps, whereas newer models incorporate object‑centric slots, 4D occupancy fields, and trajectory‑conditioned latents to capture richer semantics and multi‑agent interactions.
\item Recent approaches move beyond supervised training by adopting self‑supervised occupancy learning, lightweight multi‑scale designs, and editable BEV sketches, paving the way toward closed‑loop, generative world‑model simulation.
\vspace{-4pt} 
\end{itemize}

\subsubsection{OG-based Generation}
\label{sec:og}

\noindent\textbf{OG Representation.} An OG representation divides the driving scene into 3D voxels and assigns each cell a probability of being occupied, producing a single lattice that simultaneously tracks static structure and moving actors. This discretisation yields far richer geometric detail than 2D projections such as BEV, enabling fine-grained reasoning about scene evolution and actor interactions -- qualities that have made OGs a mainstay of future-state prediction in autonomous driving. The price of this fidelity, however, is steep: dense voxel grids demand considerable memory and computational throughput, which can hinder deployment in large-scale or real-time systems~\cite{tian2024occ3d}.

OG-based Generation (\ie, occupancy forecasting) originates from forecasting semantic occupancy grids on BEV~\cite{mahjourian2022occupancy,schreiber2019long,casas2021mp3,sadat2020perceive,toyungyernsub2022dynamics,khurana2022differentiable}. It is employed to predict how the surrounding occupancy will evolve in the near future.

\noindent\textbf{CNN-based OG Generation.} Occ4cast starts this type by unifying sparse-to-dense completion and 4D forecasting into a coherent Eulerian grid that reconstructs geometry and predicts its future\!~\cite{liu2024lidar}. Cam4DOcc then ports this paradigm from LiDAR to low-cost multi-view cameras, releasing a benchmark and network that align per-view geometry with motion cues\!~\cite{ma2024cam4docc}. Finally, PreWorld trims supervision demands by introducing a semi-supervised, two-stage pipeline that distills 2D imagery into 3D occupancy and planning priors\!~\cite{li2025semi}. These milestones trace a clear trajectory: from dense LiDAR-based geometry, through camera-centric low-cost sensing, to label-efficient world models.

\begin{table*}[t]
\centering
\caption{\small \textbf{Overview} of summary, methods and core architectures for OG-/PC-based generation (\S\ref{sec:og} and \S\ref{sec:pc-based}).}
    \small
    \resizebox{0.99\textwidth}{!}{
      \setlength\tabcolsep{3.5pt}{}
      \renewcommand\arraystretch{1.05}
\begin{tabular}{r||ccccccc}
\hline\thickhline\rowcolor{mygray}
\textbf{Summary} & \textbf{Method}& \textbf{Pub.} & \textbf{Core Architecture} & \textbf{Input Modality} \& \textbf{Control Condition} & \textbf{Output Modality} & \textbf{Training Dataset}\\
\hline\hline
\multicolumn{7}{c}{\textit{OG-based Generation}}\\
\hline
\multirow{3}{*}{\shortstack{OG-based Generation\\ CNNs}}& Occ4cast\cite{liu2024lidar}  & IROS'24 & CNNs & LiDAR & Occupancy & OCFBench\cite{liu2024lidar}\\
& Cam4DOcc\cite{ma2024cam4docc} & CVPR'24 & CNN & Image\,{+}\,Ego motion\,{+}\,GMO\,{+}\,GSO & Occupancy & OpenOccupancy\cite{wang2023openoccupancy} \\  
& PreWorld\cite{li2025semi}  & ICLR'25 & CNN & Image\,{+}\,Occupancy & Occupancy & Occ3D\cite{tian2024occ3d} \\
\cdashline{1-7}[1pt/1pt]
\multirow{11}{*}{\shortstack{OG-based Generation\\ Transformers}} & MUVO\cite{bogdoll2023muvo} & IV'23 & Transformer & Image\,{+}\,LiDAR\,{+}\,Action & Image\,{+}\,LiDAR\,{+}\,Occupancy & CARLA\cite{dosovitskiy2017carla}\\  
& OccWorld\cite{zheng2025occworld} & ECCV'24 & Transformer & Occupancy\,{+}\,Trajectory & Occupancy\,{+}\,Trajectory & Occ3D\cite{tian2024occ3d} \\
& DFIT-OccWorld\cite{zhang2024efficient} & arXiv'24 & Transformer & Occupancy\,{+}\,Image\,{+}\,Ego poses & Occupancy\,{+}\,Image\,{+}\,Trajectory & Occ3D\cite{tian2024occ3d}\,{+}\,OpenScene\cite{openscene2023}\\ 
& DriveWorld\cite{min2024driveworld} & CVPR'24 & Transformer & Image\,{+}\,Action & Occupancy\,{+}\,Action & nuScenes\cite{caesar2020nuscenes}\,{+}\,OpenScene\cite{openscene2023} \\
& Drive-OccWorld\cite{yang2025driving} & AAAI'25 & Transformer &  Image\,{+}\,Action  & Occupancy\,{+}\,Flow\,{+}\,Trajectory & nuScenes\cite{caesar2020nuscenes}\,{+}\,OpenOccupancy\cite{wang2023openoccupancy} \\
& OccLLaMA\cite{wei2024occllama} & arXiv'24 & Transformer & Occupancy\,{+}\,Trajectory\,{+}\,Text & Occupancy\,{+}\,Trajectory\,{+}\,Text &nuScenes\cite{caesar2020nuscenes}\,{+}\,OpenOccupancy\cite{wang2023openoccupancy} \\
& GaussianWorld\cite{zuo2024gaussianworld} & CVPR'25 & Transformer\,{+}\,3DGS & Gaussians\,{+}\,Image & Occupancy & nuScenes\cite{caesar2020nuscenes} \\
& RenderWorld\cite{yan2024renderworld} & ICRA'24 &  Transformer\,{+}\,3DGS & Image & Occupancy\,{+}\,Trajectory & nuScenes\cite{caesar2020nuscenes} \\
& OccProphet\cite{chen2025occprophet} & ICLR'25 &  Transformer & Image & Occupancy\,{+}\,Flow & nuScenes\cite{caesar2020nuscenes}\,{+}\,OpenOccupancy\cite{wang2023openoccupancy} \\
& Occ-LLM\cite{xu2025occ} & ICRA'25 & LLM & Image\,{+}\,Occupancy & Occupancy\,{+}\,Trajectory\,{+}\,Text & nuScenes\cite{caesar2020nuscenes}\,{+}\,Occ3D\cite{tian2024occ3d} \\
& T$^3$Former~\cite{xu2025temporal} & arXiv'25 & Transformer & Occupancy & Occupancy\,{+}\,Trajectory & nuScenes\cite{caesar2020nuscenes}\\
& AFMWM\cite{liu2025towards} & arXiv'25 & Transformer & Occupancy\,{+}\,Trajectory & Occupancy\,{+}\,Video & Occ3D\cite{tian2024occ3d}\,{+}\,KITTI360\cite{liao2022kitti}\\
& I$^2$-World~\cite{liao20252} & arXiv'25 & Transformer & Occupancy & 4D occupancy & Occ3D\cite{tian2024occ3d} \\
\cdashline{1-7}[1pt/1pt]
\multirow{5}{*}{\shortstack{OG-based Generation\\ Diffusion Models}} & DOME\cite{gu2024dome} & arXiv'24 & Diffusion Model & Occupancy\,{+}\,Action & Occupancy & nuScenes\cite{caesar2020nuscenes} \\
& OccSora\cite{wang2024occsora} & arXiv'24 & Diffusion Model & Occupancy\,{+}\,Trajectory & Occupancy & Occ3D\cite{tian2024occ3d} \\
& UniScene\cite{li2024uniscene} & arXiv'24 & Diffusion Model & BEV Layouts\,{+}\,Text & Occupancy\,{+}\,LiDAR\,{+}\,Image & OpenOccupancy\cite{wang2023openoccupancy}\,{+}\,nuScenes\cite{caesar2020nuscenes}\\
& DynamicCity\cite{bian2025dynamiccity} & ICLR'25 & Diffusion Model & Occupancy\,{+}\,Command\,{+}\,Trajectory\,{+}\,Layout &  Occupancy  & Occ3D\cite{tian2024occ3d}\,{+}\,CarlaSC\cite{wilson2022motionsc}\\
& COME\cite{shi2025come} & arXiv'25 & Diffusion Model & Occupancy\,{+}\,Pose/BEV layout & Occupancy & Occ3D\cite{tian2024occ3d} \\
\hline\hline
\multicolumn{7}{c}{\textit{PC-based Generation}}\\
\hline
\multirow{2}{*}{\shortstack{PC-based Generation\\CNNs}} & PCP\cite{mersch2022self} & CoRL'22 & CNN & LiDAR\,{+}\,Range map & LiDAR  & KITTI\cite{geiger2012we} \\
& 4DOcc\cite{khurana2023point} & CVPR'23 & CNN & LiDAR & Occupancy\,{+}\,LiDAR & KITTI\cite{geiger2012we}\,{+}\,nuScenes\cite{caesar2020nuscenes}\\
\cdashline{1-7}[1pt/1pt]
& PCPNet\cite{luo2023pcpnet} & RA-L'23 & CNN & LiDAR\,{+}\,Range map & LiDAR  & KITTI\cite{geiger2012we} \\
\multirow{2}{*}{\shortstack{PC-based Generation\\ Transformers}} & ViDAR\cite{yang2024visual} & CVPR'24 & Transformer & Image\,{+}\,Action & LiDAR & nuScenes\cite{caesar2020nuscenes}\\
& HERMES\cite{zhou2025hermes} & ICCV'25 & Transformer & Image\,{+}\,Text\,{+}\,Action & LiDAR\,{+}\,Text & nuScenes\cite{caesar2020nuscenes}\\
\cdashline{1-7}[1pt/1pt]
\multirow{3}{*}{\shortstack{PC-based Generation\\ Diffusion Models}} & LiDARGen\cite{zyrianov2022learning} & ECCV'22 & Diffusion Model & LiDAR\,{+}\,Range map  & LiDAR & KITTI-360\cite{liao2022kitti}\\
& Copilot4D\cite{zhangcopilot4d} & ICLR'24 & Diffusion Model & LiDAR & LiDAR & KITTI\cite{geiger2012we}\,{+}\,nuScenes\cite{caesar2020nuscenes}\\
& RangeLDM\cite{hu2024rangeldm} & arXiv'24 & Diffusion Model & LiDAR\,{+}\,Range map & LiDAR & KITTI-360\cite{liao2022kitti}\\
& LiDARCrafter\cite{liang2025lidarcrafter} & arXiv'25 & Diffusion Model & Text\,{+}\,Ego-centric scene graph\,{+}\,LiDAR\,{+}\,Motion & LiDAR & nuScenes\cite{caesar2020nuscenes}\\
\cdashline{1-7}[1pt/1pt]
\multirow{7}{*}{\shortstack{PC-based Generation\\ Others}} & Lidarsim\cite{manivasagam2020lidarsim} & CVPR'20 & Composition & LiDAR  & LiDAR & Internal dataset\cite{manivasagam2020lidarsim}\\
& lidarGeneration\cite{caccia2019deep} & IROS'19 & VAE\,{+}\,GAN & LiDAR\,{+}\,Point map  & LiDAR  & KITTI\cite{geiger2012we} \\
& UltraLiDAR\cite{xiong2023ultralidar} & CVPR'23 & VQ-VAE & LiDAR & LiDAR & KITTI-360\cite{liao2022kitti}\,{+}\,nuScenes\cite{caesar2020nuscenes} \\
& SPFNet\cite{weng2021inverting} & CoRL'21 & LSTMs & LiDAR\,{+}\,Range map & LiDAR & KITTI\cite{geiger2012we}\,{+}\,nuScenes\cite{caesar2020nuscenes}\\
& S2net\cite{weng2022s2net} & ECCV'22  & LSTMs & LiDAR\,{+}\,Range map & LiDAR & KITTI\cite{geiger2012we}\,{+}\,nuScenes\cite{caesar2020nuscenes}\\
& NFL\cite{huang2023neural} & ICCV'23 & Nerf & LiDAR & LiDAR  & KITTI\cite{geiger2012we} \\
& Nerf-lidar\cite{zhang2024nerf} & AAAI'24 & Nerf & LiDAR\,{+}\,Image\,{+}\,Segmentation & LiDAR  & nuScenes\cite{caesar2020nuscenes} \\
\hline
\end{tabular}
}
\label{tab:overview2}
\vspace{-3pt} 
\end{table*}

\noindent\textbf{Transformer-based OG Generation.} Many recent leading methods of this type adopt Transformer‑based core architectures. These models act as vision‑centric world models for autonomous driving and are designed to learn from pre‑collected, open‑loop benchmarks; hence they sit outside ``Interaction between Planning and Prediction''.

Transformer-based OG research starts with MUVO~\cite{bogdoll2023muvo}, which fuses camera~+~LiDAR features in a voxel-level Transformer to predict 3D occupancy, yet its deterministic rollout and label hunger limit scalability. OccWorld~\cite{zheng2025occworld} converts these voxels into autoregressive scene tokens, while DFIT-OccWorld~\cite{zhang2024efficient} decouples dynamic-static warping to cut training cost, collectively launching a generative, token-based line of work. Toward perception-control unity, DriveWorld~\cite{min2024driveworld} pre-trains a memory-augmented world model for downstream planning, and Drive-OccWorld~\cite{yang2025driving} adds trajectory-conditioned heads so the same tokens directly score driving costs. \textit{Efficiency} remains a parallel thread -- OccProphet~\cite{chen2025occprophet} refines compute, and T$^{3}$Former~\cite{xu2025temporal} sparsifies temporal attention with triplane features, achieving real-time, camera-only deployment. Meanwhile, AFMWM\cite{liu2025towards} pairs a high-ratio Swin-VAE compressor with efficient latent forecasting. I$^{2}$-World~\cite{liao20252} further pursues efficiency by decoupling tokenization into \emph{intra}- and \emph{inter}-scene stages within an encoder–decoder with a transformation-matrix control head. OccLLaMA~\cite{wei2024occllama} and Occ-LLM~\cite{xu2025occ} broaden the token vocabulary to \textit{language and action}, enabling instruction-driven occupancy reasoning across the same generative backbone. Collectively, the field has evolved from deterministic voxel Transformers through autoregressive token worlds and efficiency-focused triplane models to \textit{multimodal, instruction-aware, real-time world models}. Some studies attempt to introduce \textit{3D Gaussian Splatting (3DGS)} into this field. GaussianWorld~\cite{zuo2024gaussianworld} treats occupancy as a 4D forecasting task in Gaussian space, explicitly factorising ego motion, local dynamics and scene completion; RenderWorld~\cite{yan2024renderworld} removes all lidar by pairing a self‑supervised Img2Occ labeler with an AM‑VAE encoder and the same autoregressive backbone, yielding a compact, vision‑only pipeline. Overall, these studies trace a path from heavy deterministic voxel fusion to lightweight stochastic tokenised representations, from perception‑only forecasting to memory‑ and planning‑conditioned roll‑outs, and from single‑modality grids to language‑grounded multi‑sensor inputs; the overall trend converges on \textit{controllable, efficient, multi‑modal} world models.

\noindent\textbf{Diffusion-based OG Generation.} Diffusion‑based OG generation methods enable fine‑grained control by coupling spatio‑temporal diffusion transformers with compact 4D tokens. They further advance toward \textit{controllable, token-efficient, multi-modal} world models by: introducing trajectory-resampling control to steer future occupancy synthesis~\cite{gu2024dome}; compressing scenes into compact 4D tokens for coherent occupancy videos~\cite{wang2024occsora}; projecting 4D voxels onto HexPlanes and rolling them out with DiT for large-scale semantic forecasting~\cite{bian2025dynamiccity}; diffusing unified BEV semantic grids into joint video~+~LiDAR streams~\cite{li2024uniscene}; and disentangling ego-motion from scene evolution via a scene-centric forecasting branch whose guidance is injected through a tailored ControlNet into the occupancy world model~\cite{shi2025come}.

\noindent{\textbf{Summary.} Table~\ref{tab:overview2} assembles recent OG generative frameworks using CNNs, Transformers, and diffusion models for 4D occupancy forecasting. Several observations emerge:
\vspace{-4pt}
\begin{itemize}
\item The field migrates from heavy, deterministic voxel fusion to lightweight, stochastic tokenisation and further to Gaussian splats, reducing memory cost.
\item Occupancy world models evolve beyond perception‑only roll‑outs: memory‑augmented and trajectory‑conditioned heads embed planning cues, and language‑grounded tokens enable instruction‑driven reasoning.
\item Efficiency and controllability become central themes: dynamic/static warping, sparse triplane attention, and diffusion‑based controllable priors. This pushes OG generation toward real‑time, scalable simulation.
\vspace{-4pt}
\end{itemize}

\subsubsection{PC-based Generation}
\label{sec:pc-based}

\noindent\textbf{PC Representation.} A PC representation encodes the world as the raw 3D points returned by LiDAR, preserving fine-grained 3D details for vehicles, pedestrians, and surrounding infrastructure. This fine spatial fidelity makes PCs indispensable for occupancy forecasting, dynamic-scene modelling, and predictive reconstruction in autonomous driving. However, the sparsity and irregular sampling of LiDAR scans -- combined with real-time computational constraints -- continue to pose significant algorithmic challenges. In response, LiDAR point cloud generation, a subtask of 3D point cloud generation, has gained increasing attention to address these limitations and enhance downstream perception and planning capabilities.

Similar to OG-based generation, PC-based generation predicts future 3D point clouds from past LiDAR sweeps.

\noindent\textbf{CNN‐based PC Generation.} PCP~\cite{mersch2022self} stacks consecutive range images into a 3D spatio-temporal volume and applies 3D CNNs that fuse geometry with motion cues for consistent temporal modelling. 4DOcc~\cite{khurana2023point} discretizes historical scans into 4D occupancy grids and leverages lightweight CNN temporal modules to forecast future occupancy, lowering annotation demands while preserving scene fidelity.

\noindent\textbf{Transformer‐based PC Generation.} PCPNet~\cite{luo2023pcpnet} enriches range-image sequences by inserting self-attention layers that capture long-range dependencies, producing temporally consistent point clouds with stronger contextual cues. ViDAR~\cite{yang2024visual} frames vision-LiDAR pre-training as an autoregressive task: it embeds images and actions with a Transformer encoder and decodes future point sequences causally, providing geometry-aware priors that benefit downstream BEV generation. Building on this, HERMES~\cite{zhou2025hermes} fuses multi-view images and textual prompts into BEV embeddings through causal attention, then employs differentiable voxel rendering, constrained by motion and collision losses, to synthesize high-fidelity point clouds that respect scene dynamics.

\noindent{\textbf{Diffusion‐based PC Generation.}} LiDARGen~\cite{zyrianov2022learning} formulates LiDAR synthesis as a score‐matching diffusion process on equirectangular range images, producing physically plausible and controllable samples. Copilot4D~\cite{zhangcopilot4d} discretizes LiDAR observations with a VQ‐VAE and conducts discrete diffusion to autoregressively sample future sweeps, enabling unsupervised world‐model learning. RangeLDM~\cite{hu2024rangeldm} denoises latent representations of range‐view images to reconstruct accurate point clouds, combining latent diffusion with Hough‐voting projections. LiDARCrafter~\cite{liang2025lidarcrafter} extends LiDAR generation to dynamic 4D world modeling by parsing free-form text into an ego-centric scene graph, conditioning a tri-branch diffusion network for layout synthesis, and enforcing temporal coherence via an autoregressive warping module with ego/agent motion priors.

\noindent{\textbf{Other PC Generation Methods.}} Early GAN‐based frameworks~\cite{sallab2019lidar} support unconditional point‐cloud synthesis, while lidarGeneration~\cite{caccia2019deep} specifically mimics real‐world LiDAR statistics via adversarial training on range‐image projections. SPFNet~\cite{weng2021inverting} and S2net~\cite{weng2022s2net} employ LSTMs on sequential range maps with explicit compensation for sensor calibration. Lidarsim~\cite{manivasagam2020lidarsim} couples physical raycasting with learned noise models to generate near‐realistic scenes. UltraLiDAR~\cite{xiong2023ultralidar} uses discrete latent tokens to preserve structural and semantic consistency. Neural‐field methods (\ie, NFL~\cite{huang2023neural} and Nerf‐LiDAR~\cite{zhang2024nerf}) extend NeRF paradigms to LiDAR, enabling novel‐view synthesis under dynamic and partially observed conditions.

\noindent{\textbf{Summary.}} Table~\ref{tab:overview2} shows different modern LiDAR generation frameworks. Several observations can be drawn:
\vspace{-4pt}
\begin{itemize}
\item PC-based Generation leverages diverse architectures (\eg, Diffusion Models, GANs, VQ-VAE, and NeRF) to address LiDAR sparsity and irregularity while preserving geometric fidelity. Methods like LiDARGen and UltraLiDAR explicitly model physical sensor constraints, whereas Nerf-LiDAR and NFL integrate neural rendering to enhance novel-view synthesis under dynamic scenarios.
\item Recent advancements prioritize integrating world models and multi-modal inputs. However, most approaches focus on geometric priors of LiDAR data, with limited exploration of semantic-aware generation or perceptual consistency for downstream tasks.
\vspace{-4pt}
\end{itemize}

\subsection{Behavior Planning for Intelligent Agents}
\label{sec:methodologies_behavior_planning}

Behavior planning translates a rich, ever-changing scene understanding into a safe, comfortable, goal-directed trajectory. Most methods first sample diverse, kinematically feasible motions and then, guided by semantic predictions and tactical cues, select the most likely candidate~\cite{casas2021mp3,zeng2020dsdnet,hu2021safe}. This likelihood-based scoring fuses predicted dynamics, traffic rules, and comfort metrics, yielding a path that adapts smoothly to evolving conditions while meeting stringent safety requirements.

\subsubsection{Learning-based Planning}
\label{Learning}

Early autonomous-vehicle planners relied on rule-based heuristics, but these brittle hand-crafted pipelines struggle with the combinatorial variety of urban traffic and become hard to maintain as the rule set expands~\cite{karnchanachari2024towards,paden2016survey}. Learning-based motion planning gained traction once large-scale driving logs and fast GPUs became available~\cite{schwarting2018planning}. Modern architectures ingest multimodal sensor streams (\ie, LiDAR, RADAR, GPS signals~\cite{caltagirone2017lidar} and camera frames~\cite{bojarski2016end, richter2017safe}) to output lane-change choices~\cite{vallon2017machine, huy2013dynamic}, future state distributions~\cite{mozaffari2020deep}, reference paths~\cite{lefevre2015learning}, or direct control commands such as steering and throttle~\cite{pomerleau1989alvinn, lu2019personalized,prakash2021multi}.

Offline supervision or reinforcement learning allows these networks to keep online inference fast and to adapt to new scenes when supplied with sufficient diverse data. Compared with rule-based planning, data-driven planners handle uncertainty and interaction richness more gracefully, especially when reinforced by graph neural networks, attention mechanisms, or large language models that encode high-level traffic intent~\cite{jia2023towards, jia2021ide, jia2023hdgt}. However, three open issues remain: (i) generalisation: performance still degrades in out-of-distribution weather, geography, or traffic densities; (ii) interpretability: opaque latent policies impede validation and debugging~\cite{pomerleau1989alvinn,codevilla2019exploring,prakash2021multi}; and (iii) formal safety guarantees: current learning pipelines lack provable bounds required for certification~\cite{lu2024activead}. Addressing these gaps is central to scaling learning-based behavior planning from research prototypes to trustworthy real-world deployment.

\noindent{\textbf{RL \& MPC Planners.}} \textit{Model-free reinforcement-learning (RL)} planner~\cite{Qian2019Deep} employs a twin-delayed deep deterministic policy gradient (TD3) that maps high-level planning features to manoeuvre commands, delegating the final continuous trajectory to a downstream generator. AdaptiveDriver~\cite{vasudevan2024planning} performs planning with \textit{adaptive world models} by learning a closed-loop social world model and embedding it into an MPC planner. In contrast, AdaWM~\cite{wang2025adawm} formulates \emph{adaptive world-model RL}: under distribution shift it diagnoses world-model/policy mismatch via a divergence measure and triggers alignment-driven finetuning while execution is through a learned RL policy. \textit{Uncertainty-aware} variants~\cite{diehl2023uncertainty} inject stochastic ensembles or adaptive low-rank updates, exposing epistemic and aleatoric risk for more robust out-of-distribution reasoning, though at the price of heavier sampling and fine-tuning overheads. \textit{Agent-centric and analytic} approaches~\cite{nachkov2025dream} model each actor separately -- using attention-based latent intents or differentiable physics -- to capture nuanced social interactions, but they rely on accurate tracks or hand-coded dynamics. \textit{Latent-space} accelerators~\cite{li2025think2drive} plan entirely in a low-dimensional latent, enabling orders-of-magnitude faster roll-outs on modest hardware, yet abstract states can omit critical geometry such as traffic-light phase and undermine fine manoeuvres. \textit{Model-based RL} learns a predictive world model and optimizes the policy via imagination roll-outs, improving sample efficiency and closed-loop reliability; recent exemplars include Raw2Drive's dual-stream alignment between a privileged and a raw-sensor world model~\cite{yang2025raw2drive} and PIWM's DreamerV3-style individual world model~\cite{gao2024dream} that encodes inter-agent intentions via trajectory prediction, enabling imagination-based RL. Moreover, there has been a growing trend of integrating RL with vision language model (VLM) and vision-language-action (VLA) models in autonomous driving. VL‑SAFE employs offline, world‑model RL guided by a VLM: it labels logs with VLM safety scores, imagines safety‑aware roll‑outs, and trains an actor-critic policy~\cite{qu2025vl}. IRL‑VLA constructs a reward world model via inverse RL to replace costly simulator-based rewards, and subsequently uses the RWM to guide PPO-based RL training of a VLA policy~\cite{jiang2025irl}.

\noindent{\textbf{LLM-based Planners.}} \textit{Autoregressive multimodal transformers}~\cite{chen2024drivinggpt,bartoccioni2025vavim} treat driving as next-token prediction over interleaved image-and-action tokens, coupling perception, world modelling and control, and achieving competitive nuPlan results, but they demand high-end multi-GPU compute and still struggle with long-horizon consistency and safety certification. WoTE adds a lightweight bird's-eye-view predictor that scores sampled trajectories online, boosting safety with minimal latency, yet its fidelity is capped by BEV grid resolution and the quality of the trajectory sampler~\cite{li2025end}. SSR compresses dense BEV features into just 16 navigation-guided tokens, cutting FLOPs while maintaining accuracy, though aggressive sparsification risks missing rare but critical actors or occlusions~\cite{li2025Navigation}. DriveSim explores \textit{GPT-4V-style} models as internal simulators for language-conditioned planning, showing promising affordance reasoning but revealing frame-order errors and inconsistent imagined trajectories that limit closed-loop reliability~\cite{sreeram2024probing}. FSDrive~\cite{zeng2025FSDrive} reframes intermediate reasoning from text CoT to a spatio-temporal visual CoT: a VLM first acts as a world model to imagine a unified future frame overlaying future lanes and 3D boxes (spatial) together with an ordinary future frame (temporal), and then, conditioned on these visual CoT predictions, outputs the trajectory as an inverse-dynamics step; a unified pretraining scheme activates visual generation while preserving understanding.

\noindent{\textbf{Volume-based Planners.}} Hybrid \textit{cost-volume} methods first sample or generate a diverse set of candidate trajectories and then rank them with a composite cost that fuses hand-crafted comfort or rule-compliance terms with likelihoods predicted from learned occupancy or segmentation maps. To preserve interpretability, such planners explicitly build a cost map, pair it with a trajectory sampler, and execute the candidate with the lowest overall cost. Concretely, \textit{cost-volume} planners~\cite{zeng2019end,sadat2020perceive,casas2021mp3,phan2020covernet} compute a confidence score for each sampled trajectory from the cost volume, guiding the choice toward paths that best balance safety, efficiency, and rule compliance. This approach yields interpretable, safety-oriented plans, but its effectiveness diminishes when the sampler cannot cover the dense combinatorial space of urban traffic~\cite{zeng2020dsdnet,hu2022st,casas2021mp3,chitta2021neat,zeng2019end,sadat2020perceive}. By contrast, \textit{occupancy volume} method replaces heuristic costs with a learnable head applied to 4D occupancy forecasts, providing richer collision cues and enabling end-to-end fine-tuning, albeit at the expense of higher memory footprints and a reliance on dense occupancy supervision~\cite{yang2025driving}.

\noindent{\textbf{Other Planners.}} MILE~\cite{hu2022model} introduces a model-based imitation learning (IL) framework that leverages 3D geometry as an inductive bias to jointly learn static scenes, surrounding agents, and ego-vehicle states from high-resolution camera data. It acquires a generative world model and driving policy from offline expert logs, and employs action-conditioned rollouts to simulate diverse future scenarios. BEV-Planner~\cite{li2024ego}, a label-free baseline, directly regresses trajectories through cross-attention between ego-vehicle queries and temporal BEV features, with optional integration of ego status. UncAD~\cite{yang2025uncad} makes the BEV map itself probabilistic: a Map Uncertainty Estimation (MUE) head predicts BE BEVV map points and their Laplace parameters, then an Uncertainty-Guided Prediction \& Planning (UGPnP) stage generates multi-modal trajectories, and an Uncertainty–Collision-Aware Selection (UCAS) ranks them by collision risk and map confidence.

\begin{table*}[t]
  \centering
\caption{\small \textbf{Overview} of representative learning-based planners (\S\ref{Learning}). Please note that methods that use VLM/VLA only to assist RL training or safety scoring are listed under \textit{RL \& MPC planners}; methods where a (V)LLM acts as the planner at inference appear under \textit{LLM-based planners}.}
  \small
  \resizebox{0.99\textwidth}{!}{
    \setlength\tabcolsep{0.5pt}
    \renewcommand\arraystretch{1.05}
    \begin{tabular}{r||ccccccc}
\hline\thickhline\rowcolor{mygray}
\textbf{Summary} & \textbf{Method} & \textbf{Pub.} & \textbf{Core Architecture} & \textbf{Input Modality} \& \textbf{Control Cond.} & \textbf{Output Modality} & \textbf{Training Dataset}\\
\hline\hline
\multicolumn{7}{c}{\textit{RL \& MPC planners}}\\
\hline
Model-free RL & PFBD\cite{Qian2019Deep} & Electronics'19 & TD3 policy net & Gap\,{+}\,Speed\,{+}\,Curvature & Manoeuvre cmd. & Internal\cite{Qian2019Deep}\\
\cdashline{1-7}[1pt/1pt]
\multirow{2}{*}{\shortstack{Adaptive World Model}}
& AdaptiveDriver\cite{vasudevan2024planning}   & arXiv'24 & BehaviorNet\,{+}\,MPC  & Agent histories\,{+}\,HD-map & Trajectory & nuPlan\cite{caesar2021nuplan}\\
    & AdaWM\cite{wang2025adawm}       & ICLR'25 & Low-rank adaptive WM & State\,{+}\,reward & Trajectory & nuPlan\cite{caesar2021nuplan}\\
\cdashline{1-7}[1pt/1pt]
\multirow{1}{*}{\shortstack{Uncertainty-aware}}
    & UMBRELLA\cite{diehl2023uncertainty} & RA-L'23 & Ensemble\,{+}\,Unc.\ propag. & Vehicle state history & Trajectory & NGSIM\cite{henaff2018model}\,{+}\,CARLA\cite{carla_leaderboard_2022} \\
\cdashline{1-7}[1pt/1pt]
\multirow{1}{*}{\shortstack{Agent-centric/Analytic}}
    & Dream2Drive\cite{nachkov2025dream} & arXiv'25 & Diff. physics WM & State\,{+}\,goal & Trajectory & WOMD\cite{ettinger2021large}\\
\cdashline{1-7}[1pt/1pt]
Latent accelerator & Think2Drive\cite{li2025think2drive} & ECCV'24 & Latent-space planner & Latent WM & Trajectory & CARLA-V2\cite{carla_leaderboard_2022}\\
\cdashline{1-7}[1pt/1pt]
\multirow{2}{*}{\shortstack{Model-base RL}}
 & Raw2Drive\cite{yang2025raw2drive} & arXiv'25 & Neural planner\,{+}\,Policy RL & Multi-view cameras\,{+}\,Ego state & Low-level actions & CARLA V2\cite{carla_leaderboard_2022}\,{+}\,Bench2Drive\cite{jia2024bench2drive} \\
 & PIWM\cite{gao2024dream} & TIV'25 & Actor–critic RL & Agent histories/relations\,{+}\,Ego & Action\,{+}\,Trajectory & I-SIM\cite{zhang2022trajgen} \\
 \cdashline{1-7}[1pt/1pt]
\multirow{2}{*}{\shortstack{VLM/VLA }}
& VL-SAFE\cite{qu2025vl} & arXiv'25 & VLM\,{+}\,WM\,{+}\,offline safe RL & Image\,{+}\,Text\,{+}\,Offline expert logs & Low-level actions/Policy & Internal\cite{qu2025vl} \\
& IRL-VLA\cite{jiang2025irl} & arXiv'25 & IRL\,+\,PPO RL\,+\,VLA policy & Multi-view images/video\,{+}\,Ego state & Trajectory\,{+}\,Actions & NAVSIM v2\cite{dauner2024navsim} \\
\hline\hline
\multicolumn{7}{c}{\textit{LLM-based planners}}\\
\hline
\multirow{2}{*}{\shortstack{Autoreg.\ Transformer}}
    & DrivingGPT\cite{chen2024drivinggpt} & arXiv'24 & MM GPT-style decoder & Image\,{+}\,ego state & Action tokens & nuPlan\cite{caesar2021nuplan} \& NAVSIM\cite{dauner2024navsim} \\
    & VaViM\cite{bartoccioni2025vavim}    & arXiv'25 & Video-Gen.\ Transformer & Video\,{+}\,ego & Action tokens & OpenDV\cite{yang2024generalized}\& nuPlan\cite{caesar2021nuplan}\& nuScenes\cite{caesar2020nuscenes} \\
\cdashline{1-7}[1pt/1pt]
BEV scorer & WoTE\cite{li2025end} & arXiv'25 & BEV head\,{+}\,sampler & BEV grid\,{+}\,cand.\ traj. & Score vector & NAVSIM\cite{dauner2024navsim}\&Bench2Drive\cite{jia2024bench2drive}\\
\cdashline{1-7}[1pt/1pt]
Sparse-token WM & SSR\cite{li2025Navigation} & ICLR'25 & Sparse token WM & BEV features & Action tokens & nuScenes\cite{caesar2020nuscenes} \& CARLA\cite{dosovitskiy2017carla}\\
\cdashline{1-7}[1pt/1pt]
Vision-LLM sim. & DriveSim\cite{sreeram2024probing} & arXiv'24 & GPT-4V probe & Lang.\ cmd.\,{+}\,frames & Sim.\ trajectory & Internal\cite{sreeram2024probing}\\
\cdashline{1-7}[1pt/1pt]
Visual CoT & FSDrive\cite{zeng2025FSDrive} & arXiv'25 & VLM\,{+}\,visual CoT& Surround cameras\,{+}\,Nav/ego & Trajectory\,{+}\,future frames & nuScenes\cite{caesar2020nuscenes} \\
\hline\hline
\multicolumn{7}{c}{\textit{Volume-based planners}}\\
\hline
\multirow{6}{*}{\shortstack{Volume-based\\Hybrid Cost Volume}}
& DSDNet\cite{zeng2020dsdnet} & ECCV'20 & CNN cost-volume & Image\,{+}\,BEV map & Trajectory set & nuScenes\cite{caesar2020nuscenes} \& ATG4D\cite{zeng2020dsdnet} \& Precog\cite{rhinehart2019precog} \\
& ST-P3\cite{hu2022st}         & ECCV'22 & Spatio-temp.\ CNN & Image seq. & Trajectory set & nuScenes\cite{caesar2020nuscenes} \& CARLA\cite{dosovitskiy2017carla}\\
& MP3\cite{casas2021mp3}       & CVPR'21 & Unified map–predict–plan & Multi-modal & Trajectory set & nuScenes\cite{caesar2020nuscenes}\\
& NEAT\cite{chitta2021neat}    & ICCV'21 & Attention fields & Image\,{+}\,HD-map & Trajectory set & nuScenes\cite{caesar2020nuscenes}\\
& NMP\cite{zeng2019end}        & CVPR'19 & Neural motion planner & LiDAR\,{+}\,HD-map & Trajectory set & Internal\cite{zeng2019end}\\
& PPP\cite{sadat2020perceive}  & ECCV'20 & Semantic BEV planner & BEV semantics & Trajectory set & nuScenes\cite{caesar2020nuscenes}\\
\cdashline{1-7}[1pt/1pt]
Occupancy volume & Occ-WM\cite{yang2025driving} & AAAI'25 & 4D occ.\ WM\,{+}\,$Q$-net & 4D occupancy & Trajectory & nuScenes\cite{caesar2020nuscenes}\,{+}\, Occ3D\cite{tian2024occ3d} \\
\hline\hline
\multicolumn{7}{c}{\textit{Others}}\\
\hline
Imitation Learning & MILE\cite{hu2022model}  & NIPS'22 & Latent dynamics\,{+}\,Policy  & Image\,{+}\,ego actions & Action\,{+}\,BEV Segmentation  & CARLA expert logs\cite{dosovitskiy2017carla}\\
Ego Status  & BEV-Planner\cite{li2024ego} & CVPR'24 & Ego-query cross-attn\,{+}\,MLP & Images\,{+}\,Ego status & Trajectory & nuScenes\cite{caesar2020nuscenes}\\
Uncertainty-aware & UncAD\cite{yang2025uncad} & arXiv'25 & MUE (Laplace BEV)\,+\,UGPnP\,+\,UCAS & Multi-view images\,{+}\,Ego state & Trajectory & nuScenes\cite{caesar2020nuscenes}\\
\hline
\end{tabular}
}
\label{tab:overview3}
\end{table*}

\noindent{\textbf{Summary.}} Table~\ref{tab:overview3} shows compares major learning-based planning paradigms. Several observations can be drawn:
\vspace{-4pt}
\begin{itemize}
\item Modern planners fuse LiDAR, RADAR, GPS and vision, sample kinematic motions, and rank them with learned semantic, interaction, and rule cues, yielding adaptable trajectories beyond brittle heuristics.
\item Architectures diverge: RL and MPC planners fuse latent dynamics, uncertainty, safety; LLM transformers cast driving as token prediction; cost-volume planners rank sampled paths via learned occupancy costs.
\item Gaps remain: models falter in novel weather, policies stay opaque, safety proofs are lacking, and heavy compute blocks real-time urban use, urging work on robust, interpretable, certifiable planners.
\vspace{-4pt}
\end{itemize}

\subsubsection{Rule-based Planning}
\label{rule}

Rule-based planners endure because they guarantee safety and are easy to audit~\cite{gonzalez2015review,zhou2022review}. They treat motion as a deterministic optimisation over explicit dynamics and hand-crafted costs, yielding transparent behaviour. Four core families dominate: car-following laws (\eg, IDM), sampling planners (RRT, state lattice), model-predictive control, and potential-field navigation. Modern stacks combine these blocks for interpretability and safety, yet modular fusion can lose global optimality and fixed heuristics struggle in dense, fast-changing traffic.

\noindent{\textbf{Car-following Models.}} Physics-inspired longitudinal models translate the instantaneous gap, speed and closing rate into throttle or brake commands. The Intelligent Driver Model and its calibrated siblings capture free-flow, synchronized and stop‐and-go regimes with a single continuous formula~\cite{treiber2000congested}. Recent work samples multiple IDM roll-outs and selects the minimum-cost, safety-compliant candidate to improve comfort without losing closed-form guarantees~\cite{dauner2023parting}. Because these laws are purely algebraic, production-grade Adaptive Cruise Control (ACC) systems can execute them at tens-to-hundreds Hz frequencies. However, their inherently short prediction horizon is a key limitation, so modern planners augment analytic commands with map priors or predictive MPC to anticipate intersections and cut-in maneuvers better.

\noindent{\textbf{Sampling-based Planners.}} Sampling-based planners probe the vehicle's configuration space with a sequence of samples, accept only collision-free continuations, and then rank feasible motion sequences with handcrafted costs. They split into two sub-families that trade optimality for speed in different ways. \textit{Random sampling} methods grow a tree or roadmap toward stochastically chosen states, guaranteeing probabilistic completeness and -- when augmented with rewiring -- asymptotic optimality. Classic Rapidly-Exploring Random Trees (RRT)~\cite{lavalle1998rapidly,kuffner2000rrt} realise millisecond-level feasibility checks for single-query problems. Their optimal variants, RRT\* and RRG, refine paths online by incremental rewiring~\cite{karaman2010incremental,karaman2011sampling}. Closed-loop extensions embed vehicle dynamics for high-speed manoeuvres, as demonstrated on MIT's ``Talos'' car in the DARPA Urban Challenge~\cite{kuwata2009real,martin2009grandchallenge} and in multi-agent settings~\cite{pongpunwattana2004realtime}. \textit{Deterministic sampling} methods pre-tile the state–time space with kinematically feasible motion primitives, then apply graph search to pick the lowest-cost, collision-free sequence. Spatiotemporal lattices and their conformal highway variant capture curvature and speed limits explicitly~\cite{ziegler2009spatiotemporal,mcnaughton2011motion}. Polynomial libraries (cubic, quintic, B-spline) further reduce run-time by storing analytically smooth segments that satisfy dynamic bounds~\cite{li2015real,chu2012local}. Discretised terminal manifolds shrink search depth for time-critical street scenarios~\cite{werling2012optimal}, and the Stanford ``Junior'' entry shows how such libraries integrate into a full urban-driving stack~\cite{montemerlo2008junior}. Because the candidate set is finite, deterministic planners sacrifice global optimality for predictable timing, making them popular for production autonomous vehicle stacks.

\noindent{\textbf{Continuous Optimisation.}} Continuous optimisation planners rebuild vehicle paths in continuous space each control cycle, casting curvature and speed as variables in a quadratic or nonlinear program constrained by dynamics and obstacle corridors. Warm-starting with the previous solution yields smooth, near-optimal trajectories without discrete motion libraries. Apollo's Expectation–Maximisation planner alternates lateral smoothing with longitudinal speed optimisation, whereas Rastelli's continuous-curvature scheme minimises jerk and curvature variation along Bézier splines. Both show that convex or near-convex formulations can meet real-time demands while overcoming the sub-optimality and roughness of sampling-based schemes, though at the cost of continual optimisation~\cite{fan2018baidu,rastelli2014dynamic}.

\noindent\textbf{Artificial Potential-field Navigation.} Attractive and repulsive potentials characterise the goal as an energy well and obstacles as peaks, letting a robot follow the negative gradient in real time~\cite{khatib1978dynamic,khatib1986real}. Newton-style updates hasten escape from saddle points~\cite{ren2006modified}, and adaptive gain tuning lessens oscillations on public roads~\cite{bounini2017modified}. Blending the field with sigmoid curves limits curvature spikes in dense traffic~\cite{lu2020hybrid}, while embedding the gradient in a torque-based proportional–derivative steering loop unifies motion planning with low-level control~\cite{galceran2015toward}. Although local minima persist, coupling potential-field guidance with higher-level sampling or MPC preserves rapid collision avoidance without forfeiting global optimality~\cite{gonzalez2015review,zhou2022review}.

\noindent{\textbf{Summary.}} Table~\ref{tab:overview4} compares four major rule-based motion-planning paradigms. Several observations follow:
\vspace{-4pt}
\begin{itemize}
\item Rule-based motion planners span car-following, sampling, continuous optimisation and potential-field families, balancing update rate, smoothness and global optimality while remaining easy to audit.
\item Modern stacks fuse potential-field reflexes, discrete sampling diversity and MPC smoothing, yet modular design can miss global optima and fixed heuristics struggle under dense traffic shifts.
\vspace{-4pt}
\end{itemize}

\begin{table*}[t]
  \centering
  \caption{\small \textbf{Overview} of representative rule-based motion-planning algorithms (\S\ref{rule}).}
  \small
  \resizebox{0.99\textwidth}{!}{
    \setlength\tabcolsep{1.0pt}
    \renewcommand\arraystretch{1.05}
    \begin{tabular}{r||cccccc}
\hline\thickhline\rowcolor{mygray}
\textbf{Summary} & \textbf{Method} & \textbf{Publication} &
\textbf{Core Architecture} & \textbf{Input Modality \& Control Condition} &
\textbf{Output Modality}\\
\hline\hline
\multicolumn{6}{c}{\textit{Car-following models}}\\
\hline
\multirow{2}{*}{\shortstack{Car-following\\Models}}
 & IDM~\cite{treiber2000congested}                 & Phys.\ Rev.\ E'00 & Closed-form ordinary-differential law & Gap\,{+}\,Speed\,{+}\,Relative speed & Longitudinal acceleration\\
 & IDM Roll-out Sampling~\cite{dauner2023parting}  & CoRL'23 & Batched IDM roll-outs\,{+}\,cost ranking & Gap\,{+}\,Speed\,{+}\,Relative speed & Acceleration\,{+}\,Rollout\\
\hline\hline
\multicolumn{6}{c}{\textit{Sampling-based planners}}\\
\hline
\multirow{3}{*}{\shortstack{Sampling-based\\Random}}
 & RRT~\cite{lavalle1998rapidly}  & Tech.\ Rep.'98 & Tree search with random exploration & State\,{+}\,Map\,{+}\,Obstacles & Feasible path\\
 & RRT-Connect~\cite{kuffner2000rrt}  & ICRA'00 & Bi-directional RRT connection  & State\,{+}\,Map\,{+}\,Obstacles & Feasible path\\
 & RRT$^\ast$/RRG~\cite{karaman2010incremental,karaman2011sampling} & RSS'10/IJRR'11 & Rewiring optimal RRT for asymptotic optimality & State\,{+}\,Map\,{+}\,Obstacles & Near-optimal path\\
\cdashline{1-6}[1pt/1pt]
\multirow{3}{*}{\shortstack{Sampling-based\\Deterministic}}
 & Spatiotemporal Lattice~\cite{ziegler2009spatiotemporal} & IROS'09 & Pre-tiled spatiotemporal lattice search & Ego\,{+}\,State\,{+}\,Map & Trajectory\\
 & Conformal Lattice~\cite{mcnaughton2011motion}           & ICRA'11 & Highway-aligned lattice search & Ego\,{+}\,State\,{+}\,Map & Trajectory\\
 & Discrete Terminal Manifold~\cite{werling2012optimal}    & IJRR'12 & Polynomial trajectory library\,{+}\,Cost ranking & Ego\,{+}\,State\,{+}\,Map & Time-critical trajectory\\
\hline\hline
\multicolumn{6}{c}{\textit{Continuous optimisation}}\\
\hline
\multirow{2}{*}{\shortstack{Continuous\\Optimisation}}
 & Apollo EM Planner~\cite{fan2018baidu} & arXiv'18 & Alternating DP\,{+}\,spline-QP & Frenet reference line\,{+}\,Feasible tunnel & Smooth trajectory\\
 & Continuous-Curvature Spline~\cite{rastelli2014dynamic} & IV'14 & Bézier-spline continuous-curvature planning\,{+}\,Jerk-aware cost & Ego\,{+}\,Map & Smooth trajectory\\
\hline\hline
\multicolumn{6}{c}{\textit{Artificial potential-field navigation}}\\
\hline
\multirow{4}{*}{\shortstack{Artificial\\Potential-Field}}
 & Classic APF~\cite{khatib1978dynamic,khatib1986real} & CISM'78/IJRR'86 & Attractive-repulsive potential field & Goal pose\,{+}\,Obstacles & Steering command\\
 & Newton-PF~\cite{ren2006modified} & T-RO'06 & Potential-field with Newton-based update & Goal pose\,{+}\,Obstacles & Steering command\\
 & Hybrid Sigmoid-PF~\cite{lu2020hybrid,bounini2017modified} & Sensors'20/IV'17 & Potential field with sigmoid blending & Goal pose\,{+}\,Obstacles & Steering command\\
 & Torque-PD PF~\cite{galceran2015toward}   & IV'15 & Potential field within proportional–derivative steering & Potential-field value & Steering torque\\
\hline
    \end{tabular}
}
\label{tab:overview4}
\vspace{-4pt} 
\end{table*}

\subsubsection{Search-based Planning}
\label{appendix0}

Search-based planners discretise the vehicle's high-dimensional, continuous state-space into a graph whose edges are motion primitives -- short, pre-computed, kinematically feasible maneuvers. Classical graph-search techniques such as Dijkstra~\cite{gu2016runtime,bohren2009little} and the A* family~\cite{kammel2009team,boroujeni2017flexible} then find the minimum-cost path according to heuristics that balance safety, smoothness, and progress. Vehicle constraints are embedded in the primitive library; Hybrid-State A* extends this idea by blending continuous headings with discrete cells to satisfy non-holonomic bicycle-model dynamics~\cite{dolgov2008practical}.

The approach offers strong optimality guarantees and explicit handling of traffic rules and obstacles, but suffers from the resolution-complexity trade-off: finer grids or larger graphs improve path quality and continuity yet raise computational load exponentially, jeopardising real-time performance. Recent work tackles this bottleneck with (i) model-predictive A* that employs relaxed cost-to-go maps for long-horizon searches in milliseconds~\cite{ajanovic2018search}, (ii) adaptive fidelity models that switch between simple longitudinal or drift dynamics to match task urgency~\cite{ajanovic2019towards}, (iii) multi-heuristic search that fuses several admissible and inadmissible heuristics to accelerate exploration in cluttered layouts~\cite{adabala2023multi}, and (iv) Task-and-Motion Planning hybrids that sample primitives from multiple local models while maintaining optimality guarantees in nonlinear, unstable regimes~\cite{ajanovic2023search}. Together these advances extend search-based planning from structured highway settings to complex urban and off-road scenarios while retaining its interpretable, graph-optimal foundation.

\subsection{Interaction between Planning and Prediction}
\label{sec:methodologies_interaction}

Interaction between behaviour planning and future prediction forms a feedback loop: the agent chooses actions while forecasting how all traffic will react. Tight coupling lets it pick manoeuvres that stay both safe and efficient. Research has advanced from passive log replay to fully interactive world models that respond online to ego actions (Fig.~\ref{fig3}).

\noindent\textbf{Generative Methods and Open-loop Regime.} Recent generative approaches (\eg, DriveGAN~\cite{kim2021drivegan}, MagicDrive~\cite{gao2024magicdrive}, and DriveDreamer~\cite{wang2024drivedreamer}) narrow the sim-to-real gap by focusing on photorealism and synthesizing editable street-scene videos that substantially enrich training corpora; nevertheless, they merely replay pre-sampled futures without responding to online control inputs, thus confining research to a data-centric, open-loop paradigm that breaks the causal link between present actions and subsequent observations. TrafficGen~\cite{feng2023trafficgen} and LCTGen~\cite{tan2023language} generate traffic scenes from vector logs and natural-language prompts, respectively, while RealGen~\cite{ding2024realgen} retrieves and recombines template behaviours to create tag-specific or crash-critical scenarios. All three remain strictly open-loop, replaying fixed futures that ignore online control inputs. NAVSIM~\cite{dauner2024navsim} shares this non-reactive paradigm, seeking only to refine open-loop metrics; it cannot inject rare events or alter traffic rules, offering merely quasi-closed-loop scores. These limitations draw growing criticism, as open-loop benchmarks can overstate safety and performance when an agent's actions never influence the futures being evaluated.

\begin{figure*}[t]
\centering
\includegraphics[width=0.95\linewidth]{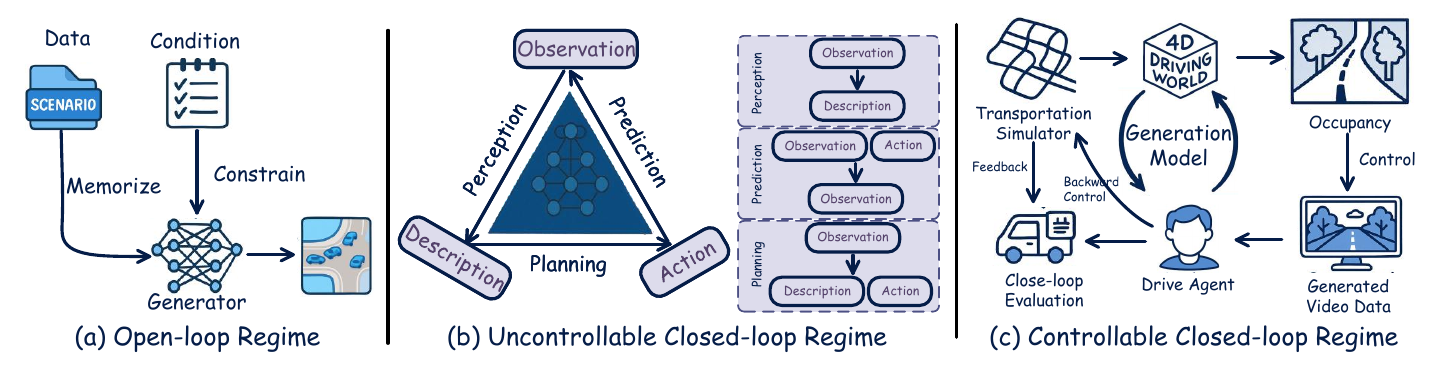}
\caption{\small \textbf{The evolution of interaction} (\S\ref{sec:methodologies_interaction}). (a) Open-loop regime synthesises scenes from logged data under static conditions; the generator memorises scenarios but never reacts to new actions. (b) Uncontrollable closed-loop regime unifies perception, prediction and planning in an auto-regressive loop, yet latent physics remain fixed, so users cannot inject rules or rare events. (c) Controllable closed-loop regime adds editable 4D worlds, occupancy control and feedback/backward signals, enabling a drive agent to interact safely with a fully testable simulator.}
\label{fig3}
\vspace{-8pt} 
\end{figure*}

\noindent\textbf{Auto-regressive World Models and Uncontrollable Closed-loop Regime.} Auto-regressive driving world models~\cite{hu2023gaia, gao2024vista, bogdoll2023muvo, zhao2024kigras, wu2024smart, yang2025resim, zheng2024doe, yang2025driving, zheng2025occworld, wei2024occllama, suo2021trafficsim,zhang2023trafficbots,zhang2025epona,feng2025Gaussian} have quickly progressed from trajectory-only roll-outs to multi-sensor, long-horizon simulators that generate futures conditioned on ego actions, unifying prediction and planning in a single generative loop. This paradigm supplies cheap, diverse data, exposes planners to their own downstream consequences, and captures uncertainty more richly than fixed-log, open-loop corpora~\cite{hu2023gaia, gao2024vista,chen2025rift}. However, the latent physics are opaque: users cannot freely edit traffic rules, inject rare events, adjust replay speed, or verify safety guarantees, and compounding distribution drift still threatens reliability, leaving current systems in an uncontrollable closed-loop that hampers rigorous safety-critical evaluation~\cite{min2024driveworld}.

Some autoregressive models rely primarily on images~\cite{hu2023gaia, gao2024vista, bogdoll2023muvo, mousakhan2025orbis}. Another line of autoregressive models~\cite{zheng2025occworld, wei2024occllama} has integrated 3D occupancy grids into environmental modeling, providing a more detailed and robust representation of spatial dynamics. These grids enable accurate handling of occlusions and spatial relationships in dense urban scenes. However, their use in long-sequence video generation remains limited by the computational complexity of processing and maintaining large-scale 3D data over extended periods. Concurrently, state-space diffusion at the agent/traffic-light level scales to trip-level city simulation\cite{tan2025scenediffuserpp}, complementing image-/occupancy-based rollouts with controllable mid-level semantics.

\noindent\textbf{Controllable Closed-loop Regime.} Research has progressed from script-driven traffic engines, through game-engine simulators with sensor realism, to neural and hybrid platforms that merge photoreal rendering with fully editable actor control. Early \textit{microscopic engines} such as SUMO~\cite{krajzewicz2012recent} exposed per-vehicle APIs for spawning, deletion and IDM/MOBIL/RL control, enabling fine-grained, reproducible experiments; Waymax~\cite{gulino2024waymax} later accelerated log replay with JAX for rapid, sensor-free policy iteration, yet neither rendered raw sensor data. To fill this gap, CARLA~\cite{Dosovitskiy17} grafted Unreal-Engine visuals and configurable sensors onto a physics core, while MetaDrive~\cite{li2022metadrive} traded photoreal texture for procedural roads and kHz physics to support fast domain randomisation. Both deliver closed-loop images but still cover limited geographies and fall short of real-world visual fidelity, restricting direct transfer to models trained on natural data. \textit{Neural-hybrid} simulators then appeared. UniSim~\cite{yang2023unisim} learns photoreal feature grids from a single drive, and OASim~\cite{yan2024oasim} re-poses vehicles with lighting-consistent implicit rendering, yet both remain bounded by their source clips. Dreamland~\cite{mo2025dreamland} links a physics simulator with a large video generator via a layered world abstraction, mapping simulator states to editable photoreal imagery and boosting controllability and realism for configurable closed-loop world creation. \textit{Fidelity and interactivity} keeps expanding: Sky-Drive~\cite{huang2025sky} adds socially aware multi-agent interactions; DriveArena~\cite{yang2024drivearena} converts 2D traffic sketches to photoreal video; DrivingSphere~\cite{yan2025drivingsphere} upgrades to 4D occupancy for geometrically accurate stress tests. The \textit{LimSim family} scales closed-loop control to city networks. LimSim~\cite{wenl2023limsim} offers distributed, scriptable traffic; LimSim++~\cite{fu2024limsim++} adds road-rule scripts and vision hooks; and the latest LIMSIM Series~\cite{fu2025limsim} provides modular APIs for covering map topology, human-style bots and area-of-interest acceleration, allowing users to inject hazards, swap policies and tune metrics for rigorous, interactive safety validation.

Beyond simulator-centric loops, recent works realize controllable closed loops \textit{inside the latent space}. World4Drive~\cite{zheng2025world4drive} builds an intention-aware latent world model, rolls out intent-conditioned futures and uses a world-model selector to choose the best trajector. LAW~\cite{li2025enhancing} learns a plan-conditioned latent world model; future-feature prediction supervises reps and waypoint planning.

\noindent\textbf{Summary.} The community has progressed from static, open-loop datasets to fully interactive, high-fidelity simulators.  Platforms that fuse controllable closed-loop dynamics with photoreal sensor generation pave the way for the next wave of behaviour-aware, safety-critical autonomous driving research. Several observations can be drawn:
\vspace{-4pt}
\begin{itemize}
    \item Closed-loop scores should complement or replace open-loop metrics to reveal true safety margins.  
    \item Generative simulators reduce reliance on real-world collection by synthesising corner cases on demand.
    \item Tight planning–prediction integration inspires hybrid curricula: models alternate between logged supervision and interactive roll-outs, and share unified losses that couple trajectory forecasting with policy optimisation.
\vspace{-4pt}
\end{itemize}

\section{Data and Training Paradigms}
\label{sec:Data_and_Training_Paradigms}
 
This section focuses on the methodologies for training models in autonomous driving, emphasizing self-supervised learning paradigms, pretraining strategies, and innovative approaches for data generation.

\begin{figure*}[t]
\centering
\includegraphics[width=0.95\linewidth]{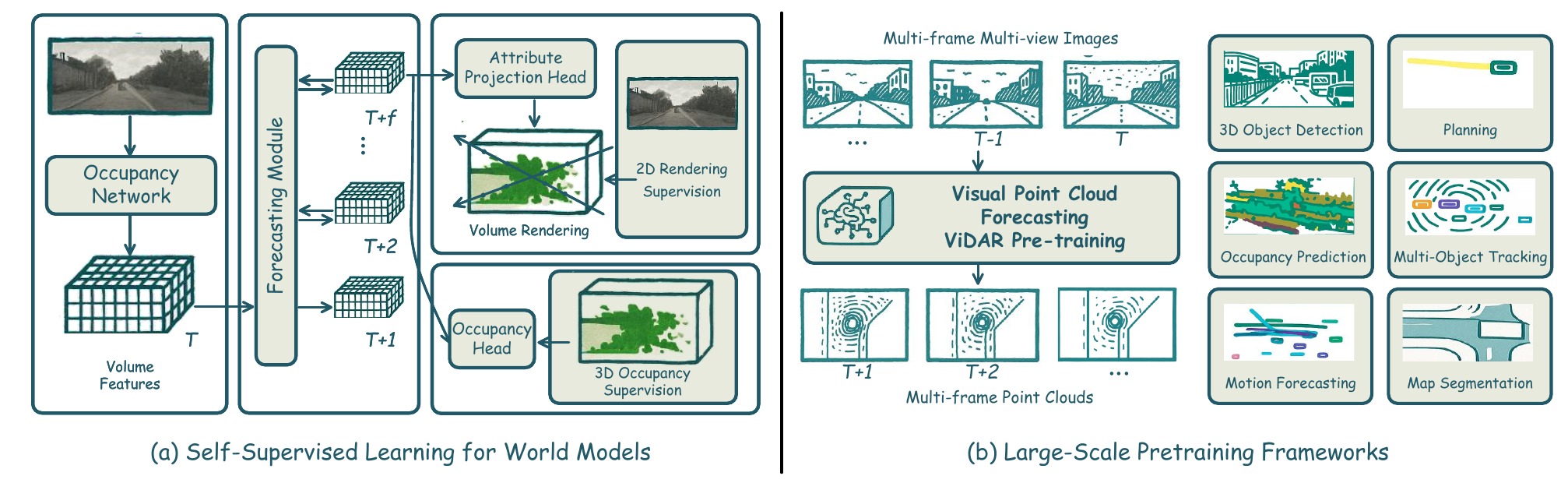}
\caption{\small (a) \textbf{Self-supervised world models} lift multi-view images into 3D volumes, forecast future grids, and learn from 2D renders plus occupancy cues, reducing labels (\S\ref{sec:Self-Supervised_Learning}). (b) \textbf{Large-scale pre-training} on visual point-cloud sequences yields a single backbone that adapts to detection, tracking, mapping, occupancy, and planning (\S\ref{sec:data_training_pretraining}).}
\label{fig4}
\vspace{-2pt} 
\end{figure*}

\subsection{Self-Supervised Learning for World Models}
\label{sec:Self-Supervised_Learning}

Self-supervised world-model research for autonomous driving uses 2D images~\cite{jin2025posepilot} or raw LiDAR scans to automatically generate supervision, cutting 3D annotation costs. Rapid progress is now reshaping the field around 3D occupancy grids. Because dense voxel annotation is laborious, many groups pivoted to \textit{cheaper} 2D supervision. RenderOcc~\cite{renderocc}, SelfOcc~\cite{huang2024selfocc}, OccNeRF~\cite{zhang2023occnerf}, H3O~\cite{shi2025h3o}, World4Drive~\cite{zheng2025world4drive}, and related NeRF-style systems lift multi-view images into volumetric space, back-propagating depth or semantic losses; these image-only routes save annotation cost but lag LiDAR-centric baselines in fidelity and fine geometry. A \textit{complementary branch} exploits the LiDAR stream itself: UnO~\cite{agro2024uno} and EO~\cite{khurana2022differentiable} derive self-labels by contrasting predicted 4D fields with future scans, while RenderWorld~\cite{yan2024renderworld}, UniPAD~\cite{yang2024unipad}, and PreWorld~\cite{li2025semi} couple differentiable rendering with multi-view cameras to fuse 2D texture and 3D structure; ViDAR~\cite{yang2024visual} further aligns video patches to point clouds for stronger temporal motion priors. \textit{Tokenisation and diffusion} bring a third flavour: COPILOT4D~\cite{zhang2023learning} and BEVWorld~\cite{zhang2024bevworld} discretise multimodal data into vocabulary codes, then learn discrete diffusion to regenerate future scenes, capturing complex distributions without any human annotation, and AD-L-JEPA~\cite{zhu2025ad} eliminates both contrastive and generative heads, directly forecasting bird's-eye embeddings for faster pre-training.

These techniques cut labeling costs, unify vision and geometry, and hint at closed-loop plan by letting a policy probe predicted futures. Current self-supervised occupancy models still face three key drawbacks. \textit{Accuracy lags behind fully-supervised 3D/4D baselines} -- NeRF-style pipelines, in particular, lose fine-scale detail. Volumetric rendering and diffusion inference incur \textit{heavy compute and memory costs}, making long-horizon prediction slow and resource-hungry. Most methods predict only the current voxel grid; \textit{principled, label-free 4D occupancy forecasting} remains largely unexplored. Looking ahead, key research directions include faster volumetric rendering, and tighter integration with downstream planners to verify that self-supervised grids truly improve safety rather than merely saving labels.

\subsection{Large-Scale Pretraining Frameworks}
\label{sec:data_training_pretraining}

A growing body of research argues that large-scale pre-training of world models is becoming the most economical path to robust autonomous driving. \textit{Universal vision/multimodal frameworks}~\cite{min2023uniworld,min2024driveworld,yang2024unipad,yang2024visual,gao2024vista,huang2024neural,zhang2024bevworld,li2025semi} ingest millions of image-LiDAR sequences, encode them in a unified BEV or voxel space, and learn to forecast future occupancies or videos. Their strengths lie in rich spatio-temporal context, easy transfer to detection, segmentation, trajectory prediction and even planning, and drastic label-cost reduction. Weaknesses include heavy GPU demand, sensitivity to camera calibration, and lingering reliance on heuristic heads (\eg, simple occupancy decoders). \textit{LiDAR/occupancy-centric self-supervision} targets pure 3D geometry. Occupancy-MAE~\cite{min2023occupancy} masks voxels, AD-L-JEPA~\cite{zhu2025ad} reconstructs masked embeddings, and UnO~\cite{agro2024uno} predicts continuous 4D occupancy fields from future point-cloud pseudo-labels. These methods excel at fine-grained geometry, learn on sparse data, and are sensor-agnostic, but they lack rich semantics and require precise temporal alignment. \textit{Foundation-level generative world models} (\eg, Gaia-1~\cite{hu2023gaia}, Gaia-2~\cite{russell2025gaia}) scale to hundreds of millions of frames, synthesize controllable multi-camera scenes, and couple generation with understanding. They promise closed-loop simulation and data augmentation, yet training cost and safety-critical evaluation remain open hurdles.

Research has progressed from early multimodal prediction networks, through \textit{unified 4D pre-training pipelines}, to the first attempts at foundation generative models; at each stage the emphasis shifts from ``perception only'' to ``perceive-predict-plan'' while annotation demand falls. Hybrid systems are anticipated that fuse dense LiDAR self-supervision with the semantic richness and controllability of vision-centric foundation models. These systems are envisioned to provide task-aware, uncertainty-calibrated priors that can be fine-tuned online. Several open challenges remain: scalable temporal memory, principled safety validation, and energy-efficient training. If these hurdles are overcome, world-model pre-training is likely to become the standard backbone for all levels of autonomous-driving software.

\subsection{Data Generation for Training}
\label{sec:data_training_generation}

Large-scale autonomous–driving datasets are still dominated by natural logs collected with expensive fleets, yet \S\ref{sec:methodologies_future_prediction} shows that modern world-model generators can manufacture much of what a learner needs. Early \textit{image-centric} systems~\cite{kim2021drivegan,wang2024drivedreamer} could only replay low-resolution clips, but the latest diffusion and Transformer models~\cite{hu2023gaia,russell2025gaia,gao2024vista,li2024drivingdiffusion} fuse text, HD-map, trajectory and sensor cues to paint photorealistic, action-conditioned street scenes that retain long-range temporal coherence. Their synthetic videos supplement rare night, rain and construction frames and feed directly into perception or closed-loop policy training. \textit{A parallel stream} converts multi-view imagery or sparse LiDAR into structured representations before generation. BEV pipelines~\cite{hu2021fiery,agro2024uno,zheng2025genad} first lift observations into a bird's-eye lattice, then roll the lattice forward to harvest scene and intention labels at centimetre accuracy while filtering away distracting appearance noise. \textit{Occupancy-grid diffusion}~\cite{wang2024occsora,bian2025dynamiccity} and \textit{tokenised voxel Transformers}~\cite{zheng2025occworld,yang2025driving} push this idea further: they stochastically sculpt 4D grids that obey physics, provide dense semantics and can be queried by planners for risk-aware cost volumes. Finally, \textit{LiDAR-native approaches}~\cite{zyrianov2022learning,hu2024rangeldm} learn score-based or latent denoisers on range images, recreating full sweeps with realistic sensor noise and occlusion artefacts; these meshes or point clouds supply geometry-heavy tasks (\eg, mapping, motion forecasting) without camera bias. Table~\ref{tab:data} summarises representative methods and how their synthetic outputs feed into training.

\begin{table}[t]
\centering
\caption{\small \textbf{Usage of data generation} (\S\ref{sec:data_training_generation}).}
\small
\resizebox{0.98\linewidth}{!}{
\setlength\tabcolsep{40pt}
\renewcommand\arraystretch{1.05}
\begin{tabular}{r||p{1.0\linewidth}}\hline\thickhline
\rowcolor{mygray} Method & Usage of data generation \\ \hline\hline
\multicolumn{2}{c}{Generation reused for downstream tasks}\\\hline
DriveDreamer-2\cite{zhao2025drivedreamer} & Generates user-customized videos from text prompts, and the authors discuss adding them to 3D detection training set.\\
BEVControl\cite{yang2023bevcontrol} & BEV sketches are converted to street-view images that fine-tune BEVFormer, boosting NDS by +1.29.\\
SimGen\cite{zhou2024simgen} & Simulator~+~real-layout images serve as augmentation, improving BEV detection \& segmentation on nuScenes.\\
GAIA-2\cite{russell2025gaia} & Synthetic multi-camera videos feed Wayve's offline perception \& planning stack for training and safety validation.\\\hline
\multicolumn{2}{c}{Generation for self-training}\\\hline
GEM\cite{hassan2024gem} & Generates scenes from ego trajectory and DINO cues.\\
Vista\cite{gao2024vista} & High-fidelity video world model evaluated on generation metrics and RL reward.\\
GAIA-1\cite{hu2023gaia} & Demonstrates controllable scene synthesis.\\
DrivingWorld\cite{hu2024drivingworld} & Video-GPT model measured by FID/FVD.\\
MiLA\cite{wang2025mila} & Targets long-horizon video synthesis for future training.\\
CoGen\cite{ji2025cogen} & Produces controllable multi-view 3D-consistent scenes.\\\hline
\end{tabular}
}
\label{tab:data}
\vspace{-6pt} 
\end{table}

Compared with hand-coded simulators, generator-based data pipelines offer three tangible benefits. 
\begin{itemize}
    \item Coverage: By conditioning on control commands, generators extrapolate to corner cases absent from logs.  
    \item Fidelity: Multi-modal diffusion preserves cross-view consistency and metre-accurate depth, which classical game engines still struggle to match.
    \item Efficiency: Once trained they amortise fleet costs, allowing continuous data refresh through cheap sampling rather than fresh drives.
\end{itemize}
However, challenges remain: high-resolution diffusion is GPU-hungry; precise camera calibration is required for view-to-view consistency; and safety assessment of purely synthetic curricula is still an open research problem. Nevertheless, data generation has evolved from a niche augmentation trick to a central pillar of autonomous-driving training pipelines, and future hybrids that blend LiDAR-tight geometry with image-level semantics are expected to yield even richer corpora for end-to-end learning.

\section{Application Areas and Tasks}
\label{sec:application}

Through the integration of world models, autonomous driving systems have made progress in tasks such as scene understanding, motion prediction, simulation, and end-to-end driving, demonstrating greater reliability and adaptability.

\subsection{Scene Understanding}
\label{sec:app_perception}

High-fidelity 3D reconstruction is vital, but real sensors are noisy and costly to label. World models answer by compressing multi-camera and LiDAR streams into a unified spatio-temporal latent that continuously refreshes as new frames arrive. \textit{Early 4D pre-training} shows that one memory state can support detection, mapping and tracking without task-specific heads~\cite{min2024driveworld}. Subsequent \textit{BEV-centric tokenisers} project all modalities into a compact bird's-eye lattice; decoding that lattice back to pixels or point clouds yields high-resolution semantics and largely removes the need for manual fusion heuristics~\cite{zhang2024bevworld}. \textit{A parallel line} predicts dense 3D occupancies instead of bounding boxes~\cite{zuo2024gaussianworld}, giving planners free-space priors that remain topologically consistent over seconds rather than frames~\cite{wang2024occsora,zheng2025occworld}. \textit{Physical realism} continues to rise as neural volumetric splats attach radiance and density to each voxel, turning the map into a differentiable oracle for visibility or risk gradients~\cite{huang2024neural,kerbl20233d}. These advances collapse sensor boundaries, cut annotation budgets, and make scene understanding an anticipatory rather than reactive module; the price is GPU memory and still-uncertain safety certificates for learned geometry.

\begin{figure*}[t]
\centering
\includegraphics[width=1.00\linewidth]{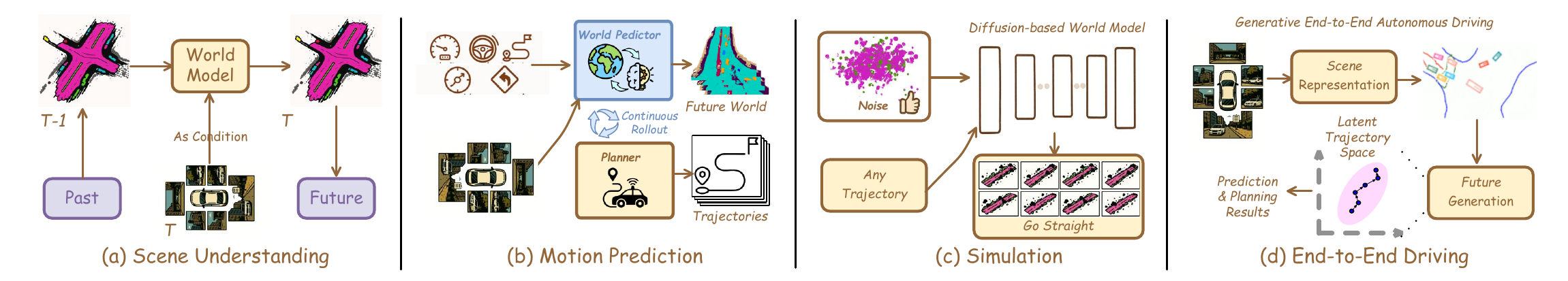}
\caption{\small \textbf{Application areas and tasks} (\S\ref{sec:application}). (a) Scene understanding fuses multi-camera/LiDAR into a 4D BEV latent that updates every frame, giving precise geometry. (b) Motion prediction rolls latent forward under action cues, outputting collision fields for planners. (c) Simulation turns the generator into an editable 4D world where diffusion seeds rare traffic for safe closed-loop tests. (d) End-to-end driving joins perception, prediction and control in one autoregressive policy.}
\label{fig5}
\vspace{-6pt} 
\end{figure*}

\subsection{Motion Prediction}
\label{sec:app_motion_prediction}

Future-state inference has moved from extrapolating agent boxes to rolling the latent world forward. Diffusion-based generators evolve the entire scene as a 4D occupancy video that enforces basic physical-consistency losses, allowing planners to clamp candidate ego trajectories and query collision risk in milliseconds~\cite{wang2024occsora}. Tokenised video models extend horizons beyond a minute while preserving fine intent cues, widening the temporal window for congestion management and logistics scheduling~\cite{guo2024infinitydrive}. Vision-centric hybrids link these 4D forecasts to an end-to-end planner, delivering action-controllable occupancy roll-outs that bypass separate mapping stages~\cite{yang2025driving}. Behaviour realism improves when multi-agent personalities are embedded inside the world; latent ``traffic bots'' yield socially plausible manoeuvres without brittle lane-change heuristics~\cite{zhang2023trafficbots,suo2021trafficsim}. The unified latent now integrates map topology, social norms, and kinematics, delivering probability fields that planners can sample or edit instead of juggling disparate trajectory sets. Remaining challenges lie in compute-heavy diffusion roll-outs and the calibration of epistemic uncertainty when imagined futures drift far from training data.

\subsection{Simulation}
\label{sec:app_simulation}

Traditional game-engine simulators cannot keep pace with the diversity and rare traffic cases. Generative world models fill that gap by producing controllable, multi-modal replicas of the road. Latent-diffusion frameworks synthesize multi-camera sequences with pixel-level realism and cross-view depth consistency, strong enough for perception-regression tests and data augmentation~\cite{hu2023gaia,gao2024vista,russell2025gaia}. Cascade pipelines further blend simulator geometry with real video priors, shrinking the sim-to-real gap and measurably improving BEV detectors when synthetic clips are mixed into training corpora~\cite{zhou2024simgen,li2024drivingdiffusion}. Geometry-first approaches generate long 4D occupancy volumes, supplying verification tools with dense physical states rather than mere imagery~\cite{guo2025dist,gu2024dome,wang2024occsora}. High-fidelity platforms now close the interaction loop, letting policies influence rendered frames each step and revealing covariate-shift failures before road deployment~\cite{yang2024drivearena,yan2025drivingsphere,yang2023unisim}. However, GPU cost and regulatory acceptance remain hurdles.

\subsection{End-to-End Driving}
\label{sec:app_e2e_driving}

End-to-end pipelines once mapped pixels directly to control signals through black-box regression; world models now provide the memory, foresight and language hooks that make such systems transparent and data-efficient. Generative architectures recast driving as next-token prediction in a multimodal language, producing future frames and waypoint plans that lower collision rates compared with deterministic baselines~\cite{zheng2025genad,chen2024drivinggpt}. Large closed-loop models integrate perception, prediction and control inside a single autoregressive policy, evaluating imagined futures every control cycle to boost rule compliance and ride comfort~\cite{zheng2024doe}. Language-vision transformers extend controllability: a single sequence model now obeys natural-language directives like ``yield at the zebra crossing, then merge'' without extra planners~\cite{chen2024drivinggpt,tan2023language}. Lightweight augmentations place a predictive head in front of imitation layers; gradients through imagined roll-outs have been reported to reduce real-world data requirements and expose intermediate occupancy maps for audit~\cite{li2025enhancing,li2025end}. Token limits and long-horizon coherence remain open issues.

\section{Performance Comparison}
\label{sec:performance}

This section presents an evaluation of world models for autonomous driving across tasks and metrics. Based on discussions in \S\ref{sec:application}, representative algorithms are benchmarked to provide evidence of their strengths and limitations.

\subsection{Evaluation Platforms and Benchmarks}
\label{appendix1}

\noindent\textbf{CarlaSC.} CarlaSC~\cite{wilson2022motionsc} is a synthetic dataset generated via CARLA, offering 4D occupancy at $128\!\times\!128\!\times\!8$ resolution within a $25.6m\!\times\!25.6m\!\times3m$ volume around the ego-vehicle.

\noindent\textbf{NuScenes.} nuScenes~\cite{caesar2020nuscenes} contains $1,\!000$ scenes split into $700$ train, $150$ val, and $150$ test. It includes six RGB cameras (360$^\circ$) and 32-beam LiDAR with centimeter-level accuracy.

\noindent\textbf{Occ3D-nuScenes.} Occ3D-nuScenes~\cite{tian2024occ3d} is built from nuScenes with $700$ training and $150$ validation sequences, each with $\sim$40 frames at $2$Hz. 

\noindent\textbf{Occ3D-Waymo.} Occ3D-Waymo~\cite{tian2024occ3d} is based on Waymo~\cite{sun2020scalability}, with $798$ training and $202$ validation sequences, each $\sim$200 frames at $10$Hz, covering $17$ semantic classes.

\noindent\textbf{OpenScene.} OpenScene~\cite{zhang2024vision}, from nuPlan~\cite{cao2022monoscene}, spans over $120$ hours across four cities with $600,\!000{+}$ frames. It uses eight cameras and five LiDARs. OpenScene-mini has $5,\!392$ train and $8,\!729$ val frames.

\subsection{4D Scene Generation}
\label{sec:perf_static_scenes}

4D Scene Generation refers to the task of synthesizing temporally coherent 3D scene sequences that evolve over time, effectively capturing both spatial and temporal dynamics.

\noindent\textbf{Metrics.} For 4D Scene Generation, both 2D and 3D perceptual quality metrics are employed: (i) Inception Score (IS)~$\uparrow$; (ii) Fr\'echet Inception Distance (FID)~$\downarrow$; (iii) Kernel Inception Distance (KID)~$\downarrow$; (iv) Precision (P)~$\uparrow$; (v) Recall (R)~$\uparrow$.

\noindent\textbf{Results.} To evaluate the effectiveness of DynamicCity~\cite{bian2025dynamiccity} in 4D scene generation, comparisons are conducted against OccSora~\cite{wang2024occsora} on the Occ3D-Waymo~\cite{tian2024occ3d} and CarlaSC~\cite{wilson2022motionsc} benchmarks. As reported in Table~\ref{tab:fid}, IS, FID, KID, P, and R are measured in both 2D and 3D spaces. Across both benchmarks and evaluation spaces, DynamicCity consistently delivers higher perceptual quality (IS, FID) and stronger fidelity-diversity trade-offs, validating its effectiveness in generating realistic 4D scenes.  Looking forward, research is converging on geometry-consistent 4D diffusion with text/trajectory/physics conditioning and unified RGB-depth-occupancy rendering to support interactive simulators.

\begin{table*}[!t]
  \centering
  \caption{\small \textbf{4D scene generation comparison} (\S\ref{sec:perf_static_scenes}). Metrics reported for OccSora~\cite{wang2024occsora} \textit{vs} DynamicCity~\cite{bian2025dynamiccity} on CarlaSC~\cite{wilson2022motionsc} and Occ3D-Waymo~\cite{tian2024occ3d}, in both 2D and 3D evaluation spaces.}
  \label{tab:fid}
  \small
\resizebox{0.99\textwidth}{!}{
\setlength\tabcolsep{13.4pt}{}
\renewcommand\arraystretch{1.05}
\begin{tabular}{r||c|c|ccccc|ccccc}
\hline\thickhline 
\rowcolor{mygray}  & &  & \multicolumn{5}{c|}{2D Metrics}         & \multicolumn{5}{c}{3D Metrics}         \\
\rowcolor{mygray}    \multirow{-2}{*}{Dataset} & \multirow{-2}{*}{Method}   & \multirow{-2}{*}{\#Frames}    & IS$\uparrow$    & FID$\downarrow$    & KID$\downarrow$    & P$\uparrow$     & R$\uparrow$     & IS$\uparrow$    & FID$\downarrow$    & 
KID$\downarrow$    & P$\uparrow$     & R$\uparrow$    \\
\hline\hline
\multirow{2}{*}{CarlaSC} & OccSora\cite{wang2024occsora} & \multirow{2}{*}{16} & 1.030 & 28.55 & 0.008 & 0.224 & 0.010 & 2.257 & 1559  & 52.72 & 0.380 & 0.151 \\
& DynamicCity\cite{bian2025dynamiccity}    &    & \textbf{1.040} & \textbf{12.94} & \textbf{0.002} & \textbf{0.307} & \textbf{0.018} & \textbf{2.331} & \textbf{354.2} & \textbf{19.10} & \textbf{0.460} & \textbf{0.170} \\
\cdashline{1-13}[1pt/1pt]
\multirow{2}{*}{Occ3D-Waymo} & OccSora\cite{wang2024occsora} & \multirow{2}{*}{16} & 1.005 & 42.53 & 0.049 & 0.654 & 0.004 & 3.129 & 3140  & 12.20 & 0.384 & 0.001 \\
 & DynamicCity\cite{bian2025dynamiccity}    &    & \textbf{1.010} & \textbf{36.73} & \textbf{0.001} & \textbf{0.705} & \textbf{0.015} & \textbf{3.206} & \textbf{1806} & \textbf{77.71} & \textbf{0.494} & \textbf{0.026} \\
\hline 
\end{tabular}
}
\end{table*}

\begin{table*}[t]
\centering
\caption{\small \textbf{Point cloud forecasting} performance on OpenScene~\cite{zhang2024vision} mini val set (\S\ref{sec:Point Cloud Forecasting}).}
\resizebox{0.99\textwidth}{!}{
\setlength\tabcolsep{34.2pt}{}
\renewcommand\arraystretch{1.05}
    \begin{tabular}{r||ccccccc} 
\hline\thickhline
\rowcolor{mygray}
\multicolumn{1}{c||}{} & \multicolumn{7}{c}{Chamfer Distance (m$^2$) $\downarrow$} \\
\rowcolor{mygray}\multicolumn{1}{c||}{\multirow{-2}{*}{Method}} &  0.5s & 1.0s & 1.5s & 2.0s & 2.5s & 3.0s & Avg.\\
\hline\hline
ViDAR~\cite{yang2024visual} & 1.34 & 1.43 & 1.51 & 1.60 & 1.71 & 1.86 & 1.58 \\
DFIT-OccWorld-O~\cite{zhang2024efficient} & 0.38 & 0.72 & 0.74 & 0.75 & 0.79 & 0.86 & 0.70 \\
DFIT-OccWorld-V~\cite{zhang2024efficient} & 0.40 & 0.75 & 0.78 & 0.83 & 0.89 & 0.90 & 0.76 \\
\hline
\end{tabular}
}
\label{tab:pc_forecast}
\vspace{-6pt} 
\end{table*}

\subsection{Point Cloud Forecasting}
\label{sec:Point Cloud Forecasting}

Point cloud forecasting is a self-supervised task that predicts future point clouds from past observations, traditionally using range-image projections processed by 3D convolutions or LSTMs while explicitly modeling sensor motion.

\noindent\textbf{Metrics.} Chamfer Distance (CD)~$\downarrow$ are employed as evaluation metric for point cloud forecasting. CD measures the similarity between a predicted point cloud and its ground-truth counterpart. Lower CD values indicate higher fidelity in capturing the true spatial distribution of points.

\noindent\textbf{Results.} Table~\ref{tab:pc_forecast} shows that both DFIT‐OccWorld variants substantially outperform the ViDAR baseline at every forecast horizon. DFIT-OccWorld-O trims mean CD to 0.70m$^2$ and stays ahead of ViDAR across all horizons; the V-variant follows at 0.76m$^2$, and the gap widens as lead time grows. The field is shifting from range-image CNN/LSTM to token-based diffusion, leveraging image priors and physics constraints to sustain long-horizon precision.

\begin{table*}[t]
\centering
\caption{\small \textbf{4D occupancy forecasting on the Occ3D-nuScenes~\cite{tian2024occ3d} dataset} (\S\ref{sec:perf_dynamic_scenes}). Aux. Sup. denotes auxiliary supervision apart from the ego trajectory. Avg. denotes the average performance of that in 1s, 2s, and 3s.}
\small
\resizebox{0.99\textwidth}{!}{
\setlength\tabcolsep{13.4pt}{}
\renewcommand\arraystretch{1.05}
\begin{tabular}{r||c|c|cccc|cccc}
\hline\thickhline \rowcolor{mygray}
&  &  &
\multicolumn{4}{c|}{mIoU $\uparrow$} & 
\multicolumn{4}{c}{IoU $\uparrow$} \\
\rowcolor{mygray} \multirow{-2}{*}{Method}  & \multirow{-2}{*}{Input} & \multirow{-2}{*}{Aux. Sup.} & 1s & 2s & 3s & Avg.  & 1s & 2s & 3s & Avg. \\
\hline\hline
Copy\&Paste & 3D-Occ & None & 14.91 & 10.54 & 8.52 & 11.33 & 24.47 & 19.77 & 17.31 & 20.52  \\
OccWorld~\cite{zheng2025occworld} & 3D-Occ & None  & 25.78 & 15.14 & 10.51 & 17.14   & 34.63 & 25.07 & 20.18 & 26.63 \\
RenderWorld~\cite{yan2024renderworld} & 3D-Occ & None  & 28.69 & 18.89 & 14.83 & 20.80 & 37.74 & 28.41 & 24.08 & 30.08\\
OccLLaMA-O~\cite{wei2024occllama} & 3D-Occ & None  & 25.05& 19.49& 15.26& 19.93& 34.56& 28.53& 24.41& 29.17\\
Occ-LLM~\cite{xu2025occ} & 3D-Occ & None & 24.02 & 21.65& 17.29& 20.99 & 36.65 & 32.14 & 28.77 & 32.52\\
DFIT-OccWorld-O~\cite{zhang2024efficient} & 3D-Occ & None & 31.68& 21.29& 15.18& 22.71& 40.28& 31.24& 25.29& 32.27\\
DOME-O~\cite{gu2024dome} & 3D-Occ & None & 35.11& 25.89 &20.29& 27.10& 43.99& 35.36& 29.74& 36.36\\
AFMWM\cite{liu2025towards} & 3D-Occ & None & 36.42 & 27.39 & 21.66 & 28.49 & 43.68 & 36.89 & 31.98 & 37.52\\
UniScene~\cite{li2024uniscene} & 3D-Occ & None & 35.37 & 29.59 & 25.08 & 31.76 & 38.34 & 32.70 & 29.09 & 34.84 \\
I$^2$-World-O~\cite{liao20252} & 3D-Occ & None & 47.62 & 38.58 & 32.98 & 39.73 & 54.29 & 49.43 & 45.69 & 49.80 \\
T$^3$Former-O~\cite{xu2025temporal} & 3D-Occ & None & 46.32 & 33.23 & 28.73 & 36.09 & 77.00 & 75.89 & 76.32 & 76.40\\
    \cdashline{1-11}[1pt/1pt]
RenderWorld~\cite{yan2024renderworld} & Camera & Occ  & 2.83 & 2.55 & 2.37 & 2.58 & 14.61 & 13.61 & 12.98 & 13.73\\
TPVFormer~\cite{huang2023tri}+Lidar+OccWorld-T~\cite{zheng2025occworld} & Camera & Semantic LiDAR  & 4.68 & 3.36 & 2.63 & 3.56 & 9.32 & 8.23 & 7.47 & 8.34 \\
TPVFormer~\cite{huang2023tri}+SelfOcc~\cite{huang2024selfocc}+OccWorld-S~\cite{zheng2025occworld} & Camera & None & 0.28 & 0.26 & 0.24 & 0.26 &  5.05 & 5.01 & 4.95 & 5.00 \\
OccWorld-F~\cite{wei2024occllama} & Camera & Occ & 8.03 & 6.91 & 3.54 & 6.16 & 23.62& 18.13& 15.22& 18.99\\
OccLLaMA-F~\cite{wei2024occllama} & Camera & Occ  & 10.34 &8.66& 6.98& 8.66& 25.81 &23.19& 19.97 &22.99\\
GWM~\cite{feng2025Gaussian} & Camera & Occ & 11.63 & 10.07& 8.17 & 10.12& 26.22& 24.97& 22.13 &24.60\\
DOME-F~\cite{gu2024dome} & Camera & None & 24.12 &17.41& 13.24 &18.25& 35.18& 27.90& 23.435 &28.84\\
DFIT-OccWorld~\cite{zhang2024efficient} & Camera&  3D-Occ & 13.38&  10.16&  7.96 & 10.50&  19.18&  16.85&  15.02 & 17.02\\
I$^2$-World-STC~\cite{liao20252} & Camera & None & 21.67 & 18.78 & 16.47 & 18.97 & 30.55 & 28.76 & 26.99 & 28.77\\
T$^3$Former-F~\cite{xu2025temporal} & Camera & Occ & 24.87 & 18.30 & 15.63 & 19.60 & 38.98 & 37.45 & 31.89 & 36.11\\ 
\hline
\end{tabular}}
\label{occforecast}
\vspace{-2pt} 
\end{table*}

\begin{table*}[t]
    \centering
\caption{\small \textbf{Motion planning on the nuScenes~\cite{caesar2020nuscenes} dataset} (\S\ref{sec:perf_planning}). Aux.Sup. denotes auxiliary supervision apart from the ego trajectory. We source the results from their respective papers.}
\small
\resizebox{0.99\textwidth}{!}{
\setlength\tabcolsep{14.9pt}{}
\renewcommand\arraystretch{1.05}
\begin{tabular}{r||c|c|cccc|cccc}
\hline\thickhline
\rowcolor{mygray}  &  &  &
\multicolumn{4}{c|}{L2 (m) $\downarrow$} &
\multicolumn{4}{c}{Collision Rate (\%) $\downarrow$}  \\
\rowcolor{mygray} \multirow{-2}{*}{Method} & \multirow{-2}{*}{Input}& \multirow{-2}{*}{Aux. Sup.} & 1s & 2s & 3s & Avg.& 1s & 2s & 3s & Avg. \\
\hline\hline
NMP~\cite{zeng2019end} & LiDAR & Box \& Motion & 0.53 & 1.25 & 2.67 & 1.48 & 0.04 & 0.12 & 0.87 & 0.34  \\
FF~\cite{hu2021safe} & LiDAR & Freespace  & 0.55 & 1.20 & 2.54 & 1.43 & 0.06 & 0.17 & 1.07 & 0.43 \\
EO~\cite{khurana2022differentiable} & LiDAR & Freespace  & 0.67 & 1.36 & 2.78 & 1.60 & 0.04 & 0.09 & 0.88 & 0.33\\
 \cdashline{1-11}[1pt/1pt]
ST-P3~\cite{hu2022st} & Camera & Map \& Box \& Depth & 1.33 & 2.11 & 2.90 & 2.11   & 0.23 & 0.62 & 1.27 & 0.71\\
UniAD~\cite{hu2023planning} & Camera & { \footnotesize Map \& Box \& Motion \& Tracklets \& Occ}  & 0.48 & 0.96 & 1.65 & 1.03 & 0.05 & 0.17 & 0.71 & 0.31 \\
UniAD+DriveWorld~\cite{min2024driveworld} & Camera & { \footnotesize Map \& Box \& Motion \& Tracklets \& Occ}  & 0.34& 0.67& 1.07& 0.69& 0.04& 0.12& 0.41& 0.19 \\
DriveDreamer~\cite{wang2024drivedreamer} & Camera & Map \& Box \& Motion & - & - & - & 0.29 & - & - & - & 0.15 \\
GenAD~\cite{zheng2025genad} & Camera & Map \& Box \& Motion & 0.36& 0.83& 1.55& 0.91& 0.06& 0.23& 1.00& 0.43\\
OccWorld-T~\cite{zheng2025occworld} & Camera & Semantic LiDAR & 0.54 & 1.36 & 2.66 & 1.52 & 0.12 & 0.40 & 1.59 & 0.70 \\
OccWorld-S~\cite{zheng2025occworld} & Camera & None & 0.67 & 1.69 & 3.13 & 1.83 & 0.19 & 1.28 & 4.59 & 2.02\\
Drive-OccWorld~\cite{yang2025driving} & Camera & None & 0.32 & 0.75 & 1.49 & 0.85 & 0.05 & 0.17 & 0.64 & 0.29 \\
ViDAR~\cite{yang2024visual} & Camera & None & - & - & - & 0.91 & - & - & - & 0.23 \\
OccWorld-F~\cite{wei2024occllama} & Camera& Occ &0.45& 1.33& 2.25& 1.34& 0.08& 0.42& 1.71& 0.73\\
OccLLaMA-F~\cite{wei2024occllama} & Camera &Occ& 0.38& 1.07& 2.15& 1.20& 0.06& 0.39& 1.65& 0.70\\
RenderWorld~\cite{yan2024renderworld} & Camera & Occ & 0.48 & 1.30 & 2.67 & 1.48 & 0.14 & 0.55 & 2.23 & 0.97  \\
DFIT-OccWorld-V~\cite{zhang2024efficient} & Camera & Occ & 0.42 & 1.14 & 2.19 & 1.25 & 0.09& 0.19 & 1.37 & 0.55\\
GWM~\cite{feng2025Gaussian} & Camera & Occ & 0.24& 1.01 & 2.05 & 1.13 & 0.07 & 0.26 & 1.45 & 0.59\\
UncAD~\cite{yang2025uncad} & Camera & None & 0.33 & 0.59 & 0.94 & 0.62 & 0.10 & 0.14 & 0.28 & 0.17\\
FSDrive~\cite{zeng2025FSDrive} & Camera & Ego status & 0.14 & 0.25 & 0.46& 0.28& 0.03& 0.06& 0.21& 0.10\\
 \cdashline{1-11}[1pt/1pt]
OccWorld~\cite{zheng2025occworld} & 3D-Occ & None & 0.43 & 1.08 & 1.99 & 1.17 & 0.07 & 0.38 & 1.35 & 0.60  \\
RenderWorld~\cite{yan2024renderworld} & 3D-Occ & None & 0.35 & 0.91 & 1.84 & 1.03 & 0.05 & 0.40 & 1.39 & 0.61  \\
OccLLaMA-O~\cite{wei2024occllama} &3D-Occ& None &0.37 &1.02 &2.03& 1.14 &0.04 &0.24 &1.20& 0.49\\
DFIT-OccWorld-O~\cite{zhang2024efficient}&  3D-Occ& None & 0.38& 0.96& 1.73& 1.02& 0.07& 0.39& 0.90 &0.45\\
T$^3$Former-O~\cite{xu2025temporal} & 3D-Occ & None & 0.32 & 0.91 & 1.76 & 1.00 & 0.08 & 0.32 &0.51 & 0.30\\
\hline
\end{tabular}
}
\label{tab:sota-plan}
\vspace{-12pt} 
\end{table*}

\subsection{4D Occupancy Forecasting}
\label{sec:perf_dynamic_scenes}

This subsection examines models' capabilities to perceive and predict dynamic scenes, focusing on how moving objects and their interactions evolve over time.

\noindent\textbf{Metrics.} The evaluation of occupancy forecasting relies on mIoU and IoUs to gauge how accurately each future frame's semantic occupancy is recovered, while placing additional emphasis on temporal accuracy and consistency across multiple time horizons (\eg, 1s, 2s, 3s).

\noindent\textbf{Results.} Table~\ref{occforecast} summarizes the 4D occupancy forecasting performance on Occ3D-nuScenes~\cite{tian2024occ3d}, where predictions for 1s, 2s, and 3s into the future are assessed via mIoU and IoU. Notably, I$^2$-World-O~\cite{liao20252} and T$^3$Former-O~\cite{xu2025temporal} attain state-of-the-art results, surpassing baseline methods. Even the purely camera-based T$^3$Former-F variant remains highly competitive, reflecting the model's robustness in scenarios without direct 3D occupancy supervision. Future advances target tokenised or diffusion backbones that accept language or trajectory prompts and camera-only pipelines that cut LiDAR cost while keeping real-time viability.

\subsection{Motion planning}
\label{sec:perf_planning}

Motion planning rapidly generates collision-free, energy-aware trajectories by considering obstacles, road layout, and surrounding vehicles, promoting safety and efficiency.

\noindent\textbf{Metrics.} The evaluation of motion planning centers on key aspects such as L2 error and collision rate. L2 error quantifies how closely a planned trajectory tracks the reference or desired path, while collision rate measures the frequency of unsafe interactions with obstacles.

\noindent\textbf{Results.} Table~\ref{tab:sota-plan} reports motion-planning results on nuScenes~\cite{caesar2020nuscenes}. For camera end-to-end planners, adding rich supervision markedly helps: \mbox{UniAD+DriveWorld} reduces the average L2 error for -33\% and the average collision rate for -39\% compared with UniAD~\cite{hu2023planning}. In the 3D-occupancy input regime, T$^3$Former-O achieves the best trade-off with 1.00m average L2 and 0.30 average collision, cutting 3s collisions by $\sim$62\% and halving the average collision rate (0.60\%$\rightarrow$0.30\%) versus OccWorld. The best performance is from FSDrive (0.28m/0.10\%), but it relies on privileged ego status supervision, making it less directly comparable to fully perception-driven planners. Overall, the table highlights that richer supervision improves accuracy and safety, while occupancy-centric models can remain competitive.

\section{Future Research Directions}
\label{sec:future_directions}

This section surveys four frontier directions -- self-supervised learning, multi-modal fusion, advanced simulation, and efficient architectures -- that aim to reduce label dependence, sharpen perception and planning, boost simulation realism, and enable resource-aware deployment.

\subsection{Self-Supervised World Models}
\label{sec:future_self_supervised}

Self-supervision has already proven that large amounts of annotated data are useful but not indispensable.

\noindent\textbf{Reducing Label Dependency.} Upcoming work is expected to couple cross-modal reconstruction (images $\leftrightarrow$ LiDAR $\leftrightarrow$ occupancy) with strong physics-based consistency losses. By letting each sensor stream synthesize or critique the others, a single model can refine depth, motion, and semantics without ever seeing a hand-drawn box. Generative objectives (\eg, long-horizon video rollout, stochastic occupancy completion, or latent LiDAR synthesis) will serve as `free' supervision signals that capture rare weather, lighting, and traffic events. Combined with lightweight on-device distillation and sparse activation routing, these techniques promise to slash annotation budgets while keeping the model small enough for automotive hardware.

\noindent\textbf{Exploring Unlabeled Data Potential.} Beyond saving labels, the frontier is to \emph{unlock the structure} hidden in the ocean of raw fleet logs. Self-supervised world models will (i) maintain a long-term memory that adapts online yet respects safety guarantees, (ii) roll imagined futures to train planners entirely in latent space, and (iii) estimate aleatoric and epistemic uncertainty on-the-fly so that risk grows when the model drifts away from familiar distributions. Coupling these world models with reinforcement learning agents turns every kilometre (real or simulated) into a self-improvement step, gradually building a driving policy that requires fewer interventions and generalises across cities.

\subsection{Multi-Modal World Models}
\label{sec:future_multimodal}

Packing images, point clouds, and BEV cues into a single latent already lifts perception, prediction, and planning. The next wave seeks universal embeddings that ingest any sensor (\eg, camera, LiDAR, radar, event, or thermal) without hand-tuned adapters. Three paths emerge: (i) ultra-compact cross-modal tokenisers that keep geometry and texture yet run on car-grade GPUs; (ii) self-aligning schedulers that weld asynchronous packets into causally consistent world states; (iii) curriculum objectives uniting differentiable rendering, language grounding, and physics-aware roll-outs so semantics, intent, and constraints co-evolve. Realising these goals would give vehicles a single, continuously refreshed memory that streamlines control, supports life-long adaptation, and sustains robustness under sensor degradation.

\subsection{Advanced Simulation}
\label{sec:advanced_simulation}

\noindent\textbf{Physics Engine.} Physics-aware generation steers autonomous-driving simulators. DrivePhysica aligns ego/world frames, injects 3D flows, and applies box-guided occlusion, yielding multi-camera videos~\cite{yang2024pysical}. Future platforms fuse diffusion world models with differentiable rigid, soft, fluid, and thermal solvers in one loop. Diffusion priors sketch 4D occupancy while the solver enforces Newtonian laws, ensuring validity and gradients that reveal traction loss, drag, or sensor occlusion before deployment.

\noindent\textbf{Cross-Scenario Generalization.} Next-generation simulators will synthesize domain-agnostic worlds spanning cities, seasons, and driving styles. Interaction corpora toughen policies in unseen regions, while text-guided generators inject rare hazards and cultural norms to push controllers beyond sensor priors. Hybrid neural-physics engines let weather, lighting, and dynamics be tuned on the fly, unifying perception tests, policy search, and safety proofs. The goal is a fully differentiable sandbox where traffic rules, multi-agent uncertainty, and physics co-evolve, enabling auto-curricula and online co-training to shorten real-world validation.

\noindent\textbf{Real-World Validation.} Modern volumetric rendering, panoramic synthesis, and cross-modal alignment let simulators mirror city traffic with centimetre geometry, photorealism across weather, and agent motion. This fidelity boosts detector pre-training, fortifies long-horizon planners, and exposes corner-case failures before road tests, while physics-based sensors and diffusion controllers keep synthetic trajectories feasible. Continuous loops align simulated outputs with fleet telemetry, producing scalable, balanced scenarios that accelerate research and validation across perception, prediction, and control.

\noindent\textbf{Diffusion-based Generation.} Diffusion-based world modelling is converging toward unified generators that synthesise kilometre-long journeys while remaining tractable on embedded hardware. Progress unfolds along four axes. First, spatial coverage widens from single images to volumetric 4D roll-outs, giving planners seconds of geometry and exposing long-horizon corner cases. Second, richer conditioning blends panoramic cameras, text, motion plans, and physics priors so each sample obeys traffic rules, weather, and social etiquette. Third, efficiency improves through tokenised latents, causal masking, and hierarchical caches that cut memory and latency without sacrificing detail. Fourth, interactive editing enables on-the-fly swaps of agents, lanes, or illumination, turning the generator into a reusable testbed. Paired with emerging perception-plus-control benchmarks, these strands point toward self-improving simulators that shorten validation cycles, reduce data costs, and hasten autonomous-driving deployment.

\subsection{Efficient World Models}
\label{sec:future_unified_models}

Looking forward, efficient world models will rely on lean, latency-aware backbones that fuse sensing, prediction, and control in a single sweep. One shared feature reservoir -- drawing jointly on vision, depth, and map cues -- will serve every downstream head, eliminating costly re-projection. Lightweight language adapters will weave route guidance and traffic regulations into this common space. Scene motion will be roughed out as coarse flow; only ambiguous regions will be up-sampled into dense volumes, slashing computation. Multi-scale grids will stream context before zooming into high-detail regions around dynamic agents, conserving memory. Capacity-elastic layers will monitor feature entropy and activate deeper blocks only when congestion rises, sustaining real-time performance on modest vehicle GPUs. Coupled with edge quantisation and federated updates, these systems will learn continually from fleet data while respecting power, bandwidth, and privacy limits -- bringing scalable, safety-critical autonomy within tangible reach.

\section{Conclusion}
\label{sec:conclusion}

World models have rapidly become a cornerstone for autonomous driving, enabling deeper integration among perception, prediction, and decision-making. Recent advances in multi-modal fusion unify data from cameras, LiDAR, and other sensors, while self-supervised learning and large-scale pretraining reduce dependence on annotated datasets. Generative methods, particularly diffusion-based approaches, now facilitate diverse synthetic data for long-tail scenarios, enhancing model robustness in rare or extreme conditions. New frameworks tightly couple motion prediction with planning algorithms, moving toward closed-loop paradigms that promise safer, more adaptive navigation. As sensing technologies evolve and cross-domain datasets proliferate, world models are poised to become even more integral to the reliable, large-scale deployment of next-generation autonomous driving systems.


\bibliographystyle{ACM-Reference-Format}
\bibliography{egbib2}










\end{document}